\documentclass[lettersize,journal]{IEEEtran}
\usepackage{amsmath,amsfonts,amssymb}
\usepackage{algorithm} 
\usepackage{algorithmic}  
\usepackage[algo2e]{algorithm2e} 
\usepackage{array}
\usepackage[caption=false,font=normalsize,labelfont=sf,textfont=sf]{subfig}
\usepackage{textcomp}
\usepackage{stfloats}
\usepackage{url}
\usepackage{caption}
\usepackage{verbatim}
\usepackage{tablefootnote}
\usepackage{multirow}
\usepackage{authblk}
\usepackage{textcomp}
\usepackage{subfig}
\usepackage{graphicx}

\usepackage{cite}

\begin{document}

\title{End-to-End triplet loss based fine-tuning for network embedding in  effective PII detection}

\author[1]{Rishika Kohli}
\author[2]{Shaifu Gupta}
\author[3]{Manoj Singh Gaur}
\affil[1,2]{Department of Computer Science and Engineering, Indian Institute of Technology Jammu, J\&K, India.}
\affil[3]{Indian Institute of Technology Jammu, J\&K, India.}


\maketitle

\begin{abstract}
There are many approaches in mobile data ecosystem that inspect network traffic generated by applications running on user's device to detect personal data exfiltration from the user's device. State-of-the-art methods rely on features extracted from HTTP requests and in this context, machine learning involves training classifiers on these features and making predictions using labelled packet traces. However, most of these methods include external feature selection before model training. Deep learning, on the other hand, typically does not require such techniques, as it can autonomously learn and identify patterns in the data without external feature extraction or selection algorithms. In this article, we propose a novel deep learning based end-to-end learning framework for prediction of exposure of personally identifiable information (PII) in mobile packets. The framework employs a pre-trained large language model (LLM) and an autoencoder to generate embedding of network packets and then uses a triplet-loss based fine-tuning method to train the model, increasing detection effectiveness using two real-world datasets. We compare our proposed detection framework with other state-of-the-art works in detecting PII leaks from user's device.
\end{abstract}


\section{Introduction}
\IEEEPARstart{A}{s} of January 2024, the Google Play Store reached a whopping 2.43 million apps, having crossed the 1 million mark in July 2013 \cite{statista2024}. With such a huge number of mobile apps available, each having different levels of security and privacy, it's crucial for users to know which apps they are using that leak their personal data. Some apps could (un)intentionally share users' personal data, making them vulnerable to social engineering attacks where individuals are manipulated into revealing valuable and sensitive information to cyber-criminals. Therefore, detection of personal data exfiltration from the user's device, respecting user's privacy becomes important. In this regard, India passed Digital Personal Data Protection Act \cite{dpdp} in 2023, which lays guidelines on the usage of personal data in a way that respects people's right to keep their information private. 

There are many approaches such as static analysis \cite{liu, androidleaks, flowdroid, lightweight, leakminer}, dynamic taint based \cite{taintdroid}, and dynamic network based \cite{recon,antsh, ants, antmo, SPCOM, mobipurpose, fedpacket, privacyguard} analysis that can be used to analyze the network traffic of mobile devices in an efficient and secure manner. In this paper, our focus is on dynamic network based approaches that inspect packets transmitted out of mobile devices so as to detect PII. This information can then be used to make user's aware of their outgoing sensitive data and take actions such as blocking these outgoing requests. 

Prediction of PII using machine learning (ML) techniques, is formulated as a binary classification problem. Different methods have been explored for predicting PII from mobile traffic. For instance, as discussed in \cite{recon}, features are extracted from HTTP requests using bag-of-words (BOW) model. Certain heuristics using term frequency inverse document frequency (tf-idf) are then employed on those features to identify relevant features. Per-domain-per-OS based classifiers are trained on these features and  predictions are made using labeled packet traces obtained through manual or automatic mobile app testing or crowd-sourcing. Further, \cite{SPCOM} used filtering techniques on features such as removing duplicate features, removing encoded features not pointing to any meaningful information or not pertaining to any PII, clubbing features based on linguistic similarities, and removing very long and short features. Then authors used decision tree and neural network for classification on these filtered features. To best of our knowledge it is seen that existing methods include external feature selection before model training. Deep Learning (DL), on the other hand, typically does not require such techniques, as it can autonomously learn and identify patterns in the dataset without external feature extraction or selection algorithms. However, the use of DL algorithms to detect PII exfiltration from network flows remains largely unexplored.

The work in this paper is divided into two broad stages. In the first stage, we identify whether PII is leaked from any app. This is formulated as a binary classification problem where we propose an end-to-end framework to process network flows from smartphones that are represented in tabular format and then pass the processed dataset to the classifier for detection. Here, Multilayer Perceptron (MLP) remains our classifier for the detection of exfiltration. The framework employs a triplet-loss based fine-tuning method to train the model, increasing detection effectiveness. In the second stage, type of PII leaked is identified by formulating multi-label classification problem based on the best architecture for binary label classification. We also propose to evaluate our framework using k-fold cross-validation technique, to demonstrate the effectiveness for both binary classification and PII type prediction tasks. We use two publicly available datasets, ReCon \cite{recon} and AntShield \cite{ants}, to compare the performance of our detection framework. 

The key contributions of this paper are:
\begin{enumerate}
\item We propose the use of large language model accompanied by autoencoder for generating network embeddings to detect PII leaks from network flows of mobile phone applications.
\item We propose the use of FT-transformer architecture \cite{fttrans} from state-of-the-art work for our objective of PII detection.
\item We conduct extensive analysis of the proposed framework on two real-world datasets to validate its performance.
\end{enumerate}

The rest of the paper is organized as follows. Section \ref{Section2} covers related work in this area and Section \ref{preliminaries} describes preliminaries. Section \ref{ProblemFormulation} presents our problem and methodology for building detection framework. Section \ref{Section5} describes the dataset used for study and covers various experiments done in this work. Section \ref{Section7} concludes the paper with future work.

\begin{table*}
    \centering
    \caption{Comparison of existing PII detection methods}
       \label{comaprison}
       \scriptsize
    \begin{tabular}{|p{2.5cm}|p{1.8cm}|p{0.8cm}|p{0.28cm}|p{3.8cm}|p{6.5cm}|} \hline
    \textbf{Ref.}& \textbf{Category} & \textbf{P / NP}\footnotemark{} & \textbf{F.S.}\footnotemark{} & \textbf{Classifier / Technique} &  \textbf{Evaluation metric}\\ \hline  
LeakMiner\cite{leakminer} & Static & - & - & Taint propagation &  Accuracy, Analysis time. \\ \hline
Liu et al. \cite{liu} & Static & NP & \text{\sffamily \checkmark} & SVM & Precision, Recall \\ \hline
AndroidLeaks\cite{androidleaks} & Static & - & - & Taint analysis & - \\ \hline
FlowDroid\cite{flowdroid}& Static & - & -& Taint analysis & Precision, recall \\ \hline 
TaintDroid\cite{taintdroid} & Dynamic taint& - & - & Taint analysis & Operation and IPC time \\ \hline
ReCon\cite{recon}& Dynamic network & NP & \text{\sffamily \checkmark} & Per-domain-per-OS DT & Correctly classified rate, Area under the curve \\ \hline
AntShield\cite{antsh} & Dynamic network & NP & \text{\sffamily \checkmark} & per-app and single DT & Accuracy, Precision, Recall, F-measure, specificity \\ \hline
Kohli et al. \cite{ants},\cite{SPCOM}& Dynamic network & NP + P & \text{\sffamily \checkmark} & DT, NN, XAI & Accuracy, Training time \\ \hline
MobiPurpose \cite{mobipurpose} & Dynamic network & NP + P & \text{\sffamily \checkmark} & SVM, Maximum Entropy, DT models for each data type& Accuracy, Precision, Recall and F1-score\\ \hline
Bakopoulou et al. \cite{fedpacket} & Dynamic network& NP & \text{\sffamily \checkmark} & Federated SVM & F1 score\\ \hline

Pan et al. \cite{panoptispy} & Static+dynamic network &- & - & MediaExtract\cite{mediaextract} tool and manual investigation & Manual checking for media content. \\ \hline
Song et al. \cite{privacyguard} & Dynamic network & - & - & Manual analysis and then derive plugins filter for string matching & Output of \cite{taintdroid} as ground-truth and performance evaluation on network performance and battery consumption. \\ \hline
Srivastava et al. \cite{privacyproxy}& Dynamic network & - & - & Genearting signatures from traffic key-value pairs & Precision, Recall, F1 score  \\ \hline
Reardon et al. \cite{50ways} & Static+dynamic network & - & - &Using side and covert channels & Comparing runtime behaviour of app with its requested permissions. \\ \hline
Sivan et al. \cite{analysis} & Dynamic network & - & - & Regular expression based search & Comparison with samples observed by agent app on user's phone \\ \hline
Wongwiwatchai et al. \cite{lightweight} & Static& NP + P & \text{\sffamily \checkmark} & NN, LR, SVM, NB, kNN, RF & Accuracy, Precision, Recall, F1 score\\ \hline
    \end{tabular}
    \footnotemark[1]{Parametric / Non-Parametric};
     \footnotemark[2]{Feature selection required?}  
     \label{comparison}
\end{table*}

\section{Related work}
\label{Section2}
Numerous studies have explored potential privacy leaks from devices to enhance user data security and privacy. These studies are categorized into \textit{static analysis, dynamic taint-based analysis,} and \textit{dynamic network-based analysis}. Table \ref{comparison} compares sensitive data detection methods across various dimensions: category, use of ML (parametric or non-parametric models), need for prior feature selection, techniques or classifiers used, and evaluation metrics. Parametric models summarize data with a fixed set of parameters, independent of the number of training examples, while non-parametric models adapt to any functional form from the training data.
\subsection{Static analysis}
Static analysis of application source code helps to identify potential behaviors, such as accessing sensitive user data\cite{leakminer}, by decompiling the app and analyzing its source code without execution. A control flow graph maps out all possible paths that might be followed within an application program from sources, where sensitive data is read or introduced, to sinks, where this data is written out or transmitted.

Liu et al. \cite{liu} developed a system to de-escalate ad library privileges using bytecode analysis. AndroidLeaks \cite{androidleaks} mapped Android API methods to permissions and detected privacy leaks using dataflow analysis. Wongwiwatchai et al. \cite{lightweight} utilized lightweight static features to develop a classification model for identifying mobile applications that transmit PII. Their approach incorporated six machine learning algorithms: Neural Network (NN), Logistic Regression (LR), Support Vector Machine (SVM), Naive Bayes (NB), k-Nearest Neighbor (kNN), and Random Forest (RF).

However, static analysis can only suggest potential privacy violations and may yield false positives, as it lacks contextual understanding and cannot observe actual runtime behaviors. Therefore, it is often complemented by dynamic analysis to validate the findings.
\subsection{Dynamic analysis}
Dynamic analysis studies an application's runtime behavior by executing it in a controlled environment. It can be further divided into two types:
\subsubsection{Taint analysis}
Taint analysis is a type of dynamic analysis to study an executing app, by marking/tainting certain pieces of data from some points in the program (\textit{Taint Sources}) as they enter the program. This tainted data is then tracked throughout the program's execution to see how it propagates and influences other data (\textit{Taint Propagation}). The goal is to identify and monitor the paths through which sensitive or untrusted data flows (\textit{Taint Sinks}), ensuring that it does not end up in insecure or unintended locations. For instance, Enck et al. \cite{taintdroid} employed this method to monitor private sensitive information on smartphones. However, taint analysis method may be inefficient and vulnerable to control flow attacks \cite{50ways}. Scaling dynamic analysis to handle thousands of apps requires automated execution and behavioral reporting, but some code paths may be missed, providing a lower bound for app behaviors without false positives.

\subsubsection{Network-based analysis}
Network-based analysis monitors network traffic during app execution. Interaction with the app can be manual or automated using tools like UI/Application Exerciser Monkey \cite{monkey}. Network traffic is captured via a proxy server and analyzed through techniques such as string matching and ML.

ReCon\cite{recon} used C4.5-based DT to detect leaks for random users in a centralized manner. They employed a BOW model for feature extraction, using certain characters as separators to identify words. Network flows/packets were represented by binary vectors indicating word presence/absence. Heuristics such as removing features with low word frequency reduced feature count, oversampling ensured inclusion of rare PII words, and tf-idf excluded common words. Per-domain-and-OS classifiers were built, with a general classifier for domains with few samples. PII values were randomized during training to prevent model reliance on them. On the other hand, AntShield \cite{antsh} performs efficient on-device analysis using a hybrid string matching-classification approach. The AntMonitor Library \cite{antmo} intercepts packets in real-time, searching for predefined strings, and then builds classifiers for unknown PII. It uses the Binary Relevance (BR) method for multi-label classification, training separate binary classifiers for each label and employing C4.5 DT models as independent classifiers in the BR framework.

Kohli et al. \cite{ants} extended ReCon's work and proposed different variations in DT and NN models for detecting PII in network traffic. Authors used explainable AI (XAI) algorithm, SHAP to provide explanations of results and re-trained best performing models using important features selected by SHAP. Kohli et al. \cite{SPCOM} further explored the use of various feature selection and filtering techniques for improving the performance of \cite{ants} framework and used XAI algorithm LIME to explain and further improve detection framework's accuracy.

To enhance privacy in PII handling, MobiPurpose \cite{mobipurpose} parses traffic request into key-value pairs, infers data types using a bootstrapping NLP approach, and identifies data collection purposes with a supervised Bayesian model. Bakopoulou et al. \cite{fedpacket} introduced a federated learning approach for mobile packet classification, allowing devices to train a global federated SVM model without sharing raw sensitive data. 

Table \ref{comparison} summarizes works that mostly use dynamic network-based methods with prior feature selection to enhance model performance. Various classifiers are employed, from traditional ML models (SVM, DT) to advanced frameworks like NN and federated learning. Common metrics include accuracy, precision, recall, and F1 score, while taint analysis uses specialized metrics like operation time (application loading, making a phone call, etc) and IPC time to measure overhead. Some studies also consider network performance and battery consumption, which is particularly relevant for mobile apps.
\begin{figure*}
    \centering
    \includegraphics[width=\textwidth,height=0.27\textwidth]{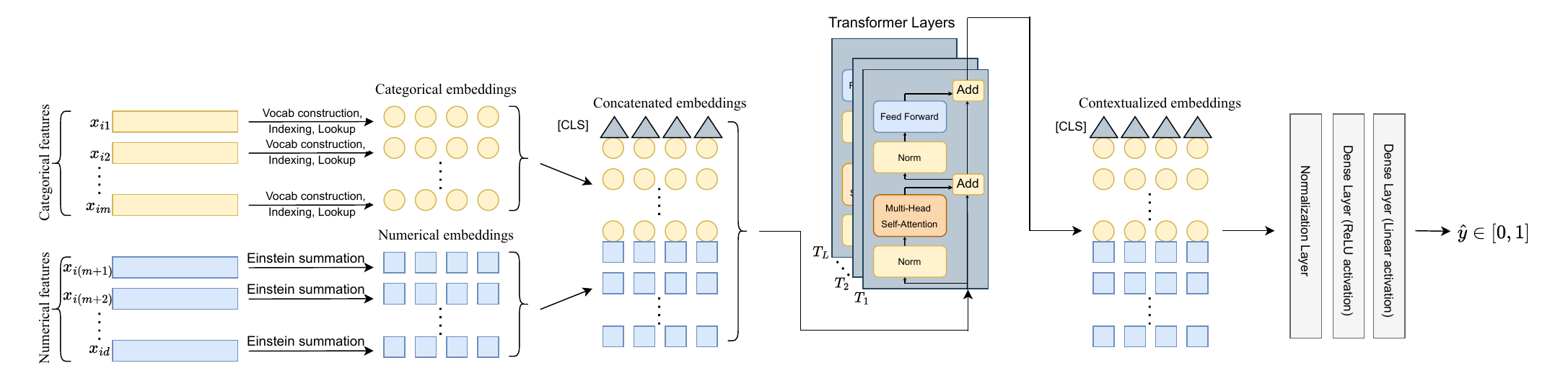}
    \caption{FT-transformer architecture}
    \label{ftransarc}
\end{figure*}

Our work falls into the category of dynamic network-based analysis, where we intercept network flows to detect PII leakage. Note that the interception of mobile traffic is not part of our contribution but is instead orthogonal to our approach. Most works in this category use ML and rely on prior feature selection strategies. With an intuition that instead of employing different feature selection techniques, can we build a model/framework that is capable enough to capture complex patterns in the input dataset without the aid of any external feature processing algorithms. Therefore, in this work we use DL and propose an end-to-end model to detect PII ex-filtration in mobile applications. In the next section, we present preliminaries and related background knowledge on the techniques used in this work.

\section{Preliminaries}
\label{preliminaries}
Before describing our proposed scheme, we first introduce a work namely FT-transformer that is our comparison benchmark. We also describe the LLM and autoencoder models used in our approach.

\subsection{FT-Transformer}
\label{fttrans-liter}
FT-Transformer \cite{fttrans} is a model that leverages a combination of a feature tokenizer and transformer components to process tabular data. The Feature Tokenizer transforms all features, both categorical and numerical, into tokens. The architecture of FT-Transformer is shown in Figure \ref{ftransarc}. Numerical features are transformed into a higher-dimensional embedding space ($ embs_{bNQ}$) using Einstein summation convention to perform matrix multiplication between the weights and the input data. 

\begin{equation*}
\begin{split}
    embs_{bNQ} &= ReLU \Bigg(\sum_{n=1}^\mathcal{N} W_{n\beta Q} \cdot x_{kn} + b_{n\beta} \Bigg)\\
    &= ReLU \Bigg(\sum_{n=1}^\mathcal{N} W_{n1Q} \cdot x_{kn} + b_{n1} \Bigg) (\because \beta=1)
    \end{split}
\end{equation*}
where, $W \in \mathbb{R}^{\mathcal{N} \times \beta \times Q}$ is the weight matrix, $\mathcal{N}$ is the number of numerical features, $\beta$ is the number of bins (=1), and $Q$ is the embedding dimension (=32). $x \in \mathbb{R}^{k \times \mathcal{N}}$ is the input matrix, $k$ is the batch size, and $b \in \mathbb{R}^{\mathcal{N} \times Q}$ is the bias. Here, each feature $n$ of the input $x_{kn}$ is scaled by the corresponding weight $W_{n1Q}$ and summed to form the embedding vector for each batch $k$.

In the case of categorical features, the input matrix $x \in \mathbb{R}^{k \times \mathcal{C}}$, where $C$ is the number of categorical features, is converted to embedding using the following steps: 
i) build a vocabulary $V_c$ using unique values in feature $c$ of input $x \in  \mathbb{R}^{k \times \mathcal{C}}$ and then convert categorical string values in $c$ to unique integer indices based on this vocabulary using string lookup, i.e., 
\begin{equation*}
    \vec{I}_c = \text{lookup}_c(x[:,c])
\end{equation*}
where $\vec{I}_c$ is a vector of integer indices for feature $c$, 
ii) the integer indices are converted to dense vectors of fixed size with $|V_c|$, vocabulary size of feature $c$ as input shape and $Q$, the embedding dimension (=32) as the output shape, i.e., 
\begin{equation*}
    \text{embs}_c(\vec{I}_c) = \phi_c[\vec{I}_c]
\end{equation*}
All feature embeddings are stacked to create:
\begin{equation*}
     \text{embs}_{bCQ} = \text{stack}([\text{embs}_c(\vec{I}_c) \text{ for } c \text{ in features}], \text{axis}=1)
\end{equation*}

The embeddings are concatenated into $E$, i.e.,  
\begin{equation*}
    E = \text{embs}_{bnq} + \text{embs}_{bcq}
\end{equation*}
and a classification token `[CLS]' is appended before passing through the $L$ Transformer layers. [CLS] token helps to aggregate information from all features, providing a comprehensive representation that can be used for accurate classification. The transformed embeddings are passed to an MLP with the first layer as normalization (LayerNorm) to facilitate optimization and enhance performance. The transformed embeddings are called contextualized embeddings as they are dynamically generated by integrating information from the entire input sequence, providing a context-aware representation for each input token.The predicted output is represented as 
\begin{equation*}
    \hat{y} = \text{Linear}(\text{ReLU}(\text{LayerNorm}(L[\text{CLS+$E$}])))
\end{equation*}
where Linear and ReLU are the activation functions used. 



\subsection{LLM model: SBert}
Bidirectional Encoder Representations from Transformers (BERT) \cite{bert} is a transformer-based language model that gained traction for generating word embeddings, enabling comparison of words based on similarity using metrics like euclidean or cosine distance. However, the original BERT model constructs embeddings solely at the word level. Thus, SBert \cite{sbert} emerged to derive independent sentence embeddings.
\subsubsection{BERT}
BERT comprises of a variable number of encoder layers and self-attention heads. Unlike traditional transformer models that incorporate both encoder and decoder mechanisms, BERT focuses solely on the encoder for language modeling tasks. Sequential models process text from either left-to-right or right-to-left but can't do both simultaneously, but the BERT processes the entire sequence of words all at once, making it bidirectional. This capability enables the model to understand the context of a word based on its entire surrounding context i.e., considering both the preceding and succeeding words to fully understand the context of each word. There are two available variants of BERT: i) BERT-Base: has 12 layers (transformer blocks), 12 attention heads, and 110 million parameters; ii) BERT-Large: comprises 24 layers , 16 attention heads, and 340 million parameters.
\subsubsection{SBERT}
Sentence Embeddings using Siamese BERT-Networks (SBert) is an adaptation of the pretrained BERT network employing siamese and triplet network architectures to extract semantically meaningful sentence embeddings, which can be compared using cosine similarity. The siamese network structure allows for the extraction of fixed-sized vectors representing input sentences. SBert incorporates a pooling operation (mean or max) on the output of BERT to obtain fixed-sized sentence embeddings.

Several pretrained sentence transformer models are available for public use, extensively evaluated for their ability to embed sentences effectively. One notable model is all-MiniLM-L6-v2, trained on a vast amount of data (over 1 billion training pairs) and designed for general-purpose use \cite{sbert-site}. It utilizes the pretrained MiniLM-L6-H384-uncased model \cite{uncased}, featuring 12 layers, 384 hidden units, 12 attention heads, and 33 million parameters, offering a speedup of 2.7x compared to BERT-Base. all-MiniLM-L6-v2 employs the BERT tokenizer with a maximum sequence length of 512 and is distilled from MiniLM-L6-H384-uncased, featuring 6 transformer layers, 12 attention heads, and a dropout probability of 10\%. Dimensionality of dense layers is 1536, utilizing the GELU (Gaussian Error Linear Unit) \cite{gelu} activation function, which weights the input based on its probability under a Gaussian distribution.

\begin{equation*}
    GELU(x)= x \cdot P(X \leq x) = x \cdot \Phi(x)
\end{equation*}
where $x$ is input to the activation function, $\Phi(x)$ is the cumulative distribution function (CDF) of the standard normal distribution. The model uses mean pooling, mapping sentences and paragraphs to a 384-dimensional dense vector space.

Therefore, SBert models can be used utilized for generating embeddings by leveraging pre-trained transformer models to encode sentences into fixed-length vectors that capture semantic meanings effectively.

\subsection{Autoencoder}
An autoencoder is a type of neural network architecture used in unsupervised learning domain that learns to compress and effectively represent input data without specific labels. Autoencoders follows two step architecture: an encoder $\alpha(\cdot)$ function that transforms the input data $E$ into a reduced latent representation $\tilde{E}=\alpha(E)$ (also called bottleneck representation). From $\tilde{E}$, a decoder $\beta(\cdot)$ function rebuilds the initial input as $\beta(\tilde{E})=E'$ and $E \approx E'$. Therefore, mathematically, the autoencoder can be represented as minimizing the reconstruction loss (usually mean squared error):
\begin{equation*}
    \mathcal{L}(E, \beta(\alpha(E))) = \frac{1}{N} \sum_{i=1}^N ||e_i -\beta(\alpha(e_i)||^2
\end{equation*}
where $e \in E$, $\alpha$ and $\beta$ are non-linear functions representing the encoder and decoder, respectively. Autoencoders can also be considered a dimensionality reduction technique, which compared to traditional techniques such as principal component analysis, can make use of non-linear transformations to project data in a lower dimensional space. 
\begin{figure*}
    \centering
    \includegraphics[width=0.9\textwidth, height=0.35\textwidth]{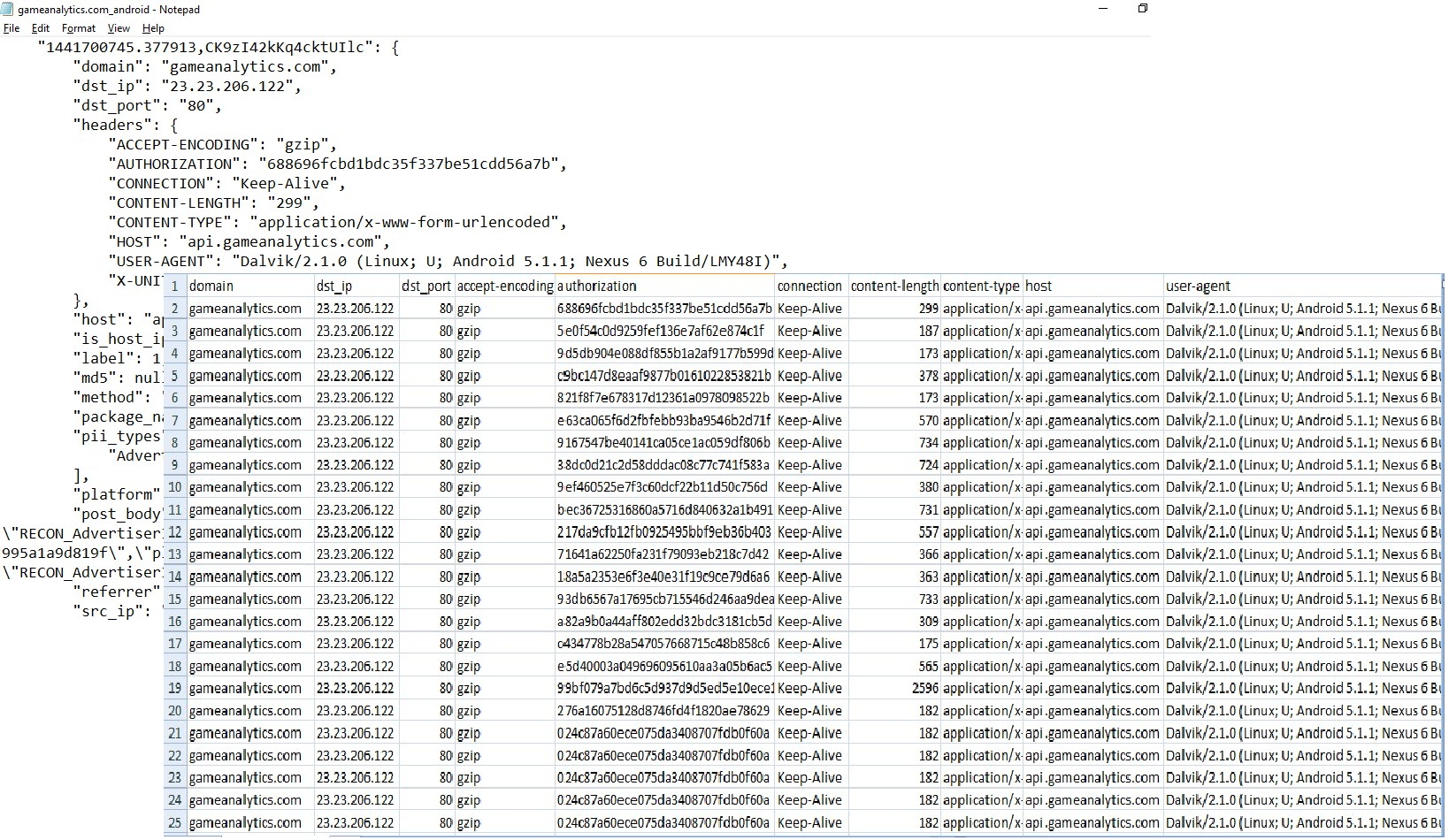}
    \caption{The pre-processed tabular dataset.}
    \label{fig1}
\end{figure*}
\section{Problem Formulation}
\label{ProblemFormulation}
In this section, we present the problem
definition and design goals, respectively.
\subsection{Problem Definition}
Smartphones can access extensive personal and sensitive information from users, which is referred to as personally identifiable information (PII) in this work. Works \cite{recon} and \cite{antsh} define a set of PII that we group into 4 categories: \textit{device identifiers, SIM card identifiers, user identifiers,} and \textit{location information}, as outlined in Table \ref{piitypes}. A network flow is deemed to transmit PII if it includes any of these types. This transmission may be: (i) to collect user information; (ii) benign, such as necessary for app functionality or acceptable to the user; or (iii) of the honest-but-curious nature. For this paper, we use ``privacy exflitration" and ``privacy leak" interchangeably.
\begin{table}
    \caption{PII List}
    \scriptsize
    \begin{tabular}{p{1.5cm}|p{6cm}}\hline
        Category &  Types\\ \hline
         Device & Advertiser ID, Android ID, IMEI, MAC Address, Device Serial Number, IDFA, MEID, X-WP-Anid\\
         SIM Card & ICCID, IMSI, Phone Number  \\
         User & Email, Name, First Name, Last Name, Gender, Password, User name, Contact name, Date-of-birth  \\ 
         Location & City, Location, Zip-code\\\hline
    \end{tabular}
    \vspace{-0.5cm}
    \label{piitypes}
\end{table}

\subsection{Design of proposed framework}
\label{Section4}
Based on our observations, it is seen that sensitive data exfiltration from mobile apps mostly occurs in structured format (i.e., key/value pairs) \cite{recon}, and more than 80\% of apps have structured responses\cite{privacyproxy , mobipurpose}, where the network flows pertaining to mobile applications can be represented in a tabular format. We broke the network flows into a tabular dataset where each key value represents a feature $f$ and value represents the sample value for that feature as shown in Figure \ref{fig1}. The problem in-hand is a classification problem having data-set $D=\{X_{ij}, y_i\}_{i=1}^N, \quad \text{for } j=1 \text{ to } d$, where $x_{i} \in 	\mathbb{R}^d$ is the $i^{th}$ data point of feature $j$, $y_i \in Y$ is the $i^{th}$ target label and $N$ is the total number of data points.  The work in this paper is divided into two stages: a) \textit{PII detection}: the target space lies in two bins, where $Y=\{0,1\}$, here 0 represents flow without PII and 1 containing PII. b)\textit{ PII classification} : target space is $Y=\{C_1,C_2,C_3,....,C_n\}$, where $C_i$ represents a specific type of PII found in network flows.

The feature set consist of $C$ categorical and $\mathcal{N}$ numerical features. The dataset now can be represented as $(X,y)$ where $X=\{X_{C},X_{\mathcal{N}}\}$ and $X_{C}$ represent samples with categorical data types and $X_{\mathcal{N}}$ are numerical data types. For instance, in Figure \ref{fig1} `domain' is the feature with samples having categorical data type and `dst\_port' has samples with numerical data type. Let $\{\vec{x}_{i1}, \vec{x}_{i2},.....\vec{x}_{im}\}\in X_{C}$ and $\{\vec{x}_{i(m+1)}, \vec{x}_{i(m+2)},.....\vec{x}_{id}\}\in X_{\mathcal{N}}$, for $i \in \{1,2,....,N\}$.

\subsubsection{PII detection using FT-transformer}
\label{fttransmethod}
As explained in Section \ref{fttrans-liter}, we first generate embedding for all $m$ categorical features i.e., $\{X_{ij}\}_{j=1}^m$ using keras embedding class `keras.layers.Embedding'. 
Let $\phi : X_{ij} \rightarrow \omega_{\phi}(X_{ij})$ for $i \in \{1,2,....,N\}$ and $j \in \{1,2,....,m\}$ be the embedding for $X_{ij}$ and $E_\phi(X_{C})$ = $ \omega_{\phi}(\vec{x}_{i1}), \omega_{\phi}(\vec{x}_{i2}),....., \omega_{\phi}(\vec{x}_{im}))$ is the set of embedding for $X_{C}$. For generating embedding for numerical features i.e., $\{X_{ij}\}_{j=m+1}^d$ by performing a linear transformation on the $X_{ij}$, followed by a rectified linear unit (ReLU) activation function. Let $\delta : X_{ij} \rightarrow \omega_{\delta}(X_{ij})$ for $i \in \{1,2,....,N\}$ and $j \in \{(m+1),(m+2),....,d\}$ be the embedding for $x_{ij}$ and $E_\delta(X_{\mathcal{N}})$ = $ \omega_{\delta}(\vec{x}_{i(m+1)}), \omega_{\delta}(\vec{x}_{i(m+2)}),....., \omega_{\delta}(\vec{x}_{id}))$ is the set of embedding for $X_{\mathcal{N}}$. The embedding $E_{FT} = \{E_\phi(X_{C})\cup E_\delta(X_{\mathcal{N}})\}$

\begin{figure*}
    \centering    
    \includegraphics[clip, trim=5cm 18cm 13cm 0cm,width=\linewidth, height=0.5\linewidth]{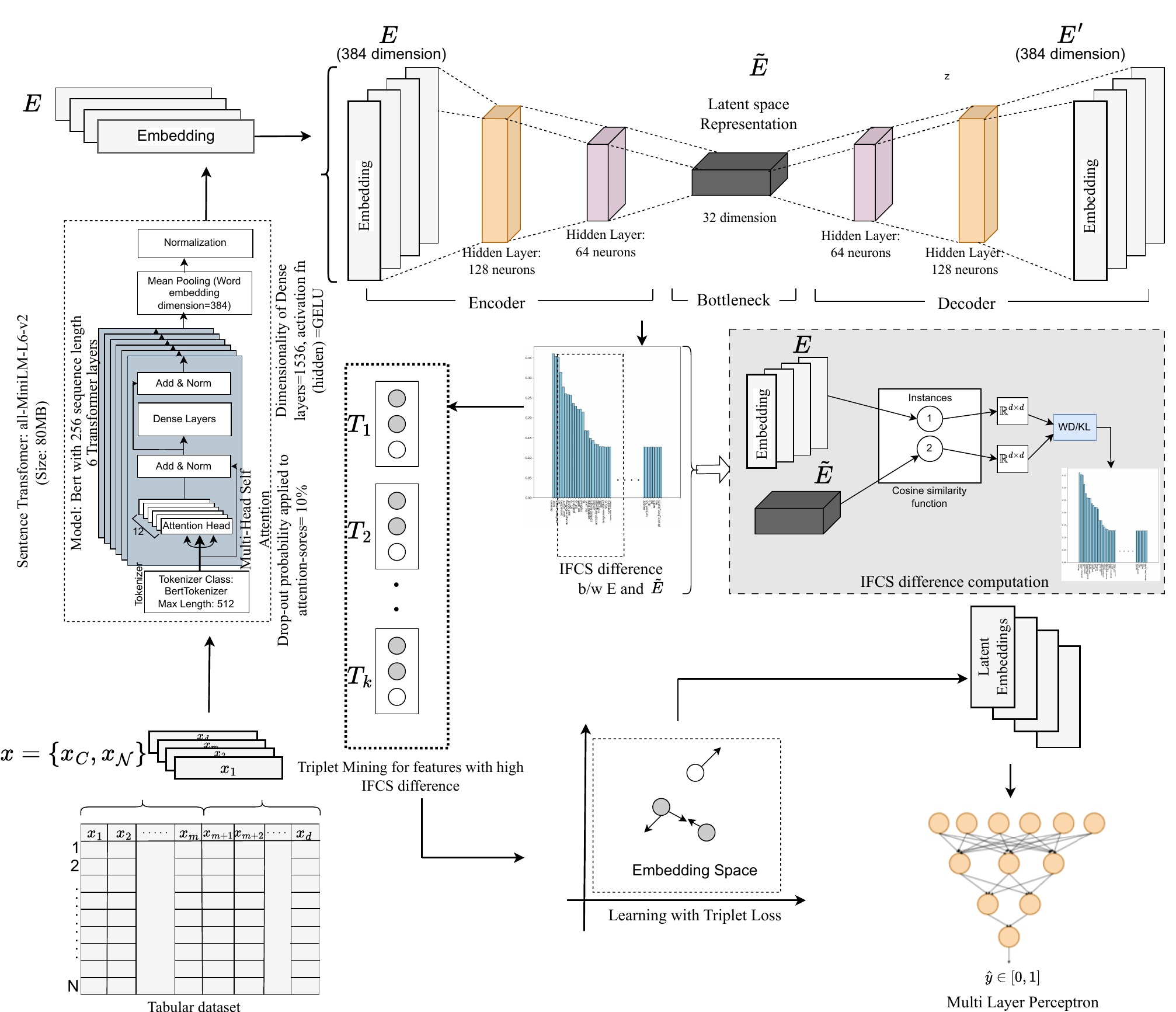}
    \caption{End-to-end learning framework}
    \label{fig-block}
\end{figure*}

Next, using FT-Transformer \cite{fttrans} architecture, the embedding $E_{FT}$ together with $CLS$ token are inputted to the first transformer layer. The output of transformer layer $L_1$ is passed to transformer layer $L_2$ and so forth. Each $E_{FT}$ is converted to contextualized embedding $E'_{FT}$ after being passed to all transformer layers. $E'_{FT}$ is then passed to MLP to predict output $\hat{y} \in [0,1]$.

\subsubsection{PII detection using LLM}
\label{PII_detection}
Next, we opted to utilize pre-trained transformers instead of initializing and training a transformer model from scratch, as done in \ref{fttransmethod}. This approach, known as transfer learning, facilitated the generation of embeddings denoted as $E$ for all features $X_{C}$ and $X_{\mathcal{N}}$. Figure \ref{fig-block} shows our end-to-end learning framework. However, the dimensions of each embedding within $E$, generated by these pre-trained transformers, are large, demanding significant computational resources for subsequent processing. To address this computational burden and learn condensed representations $\tilde{E}$ of $E$, we employed an autoencoder. We calculated cosine similarity between each feature in $E$ and in $\tilde{E}$ separately, followed by calculating the Wasserstein Distance (WD) and KL-Divergence (KL) on the similarity scores to measure the inter-feature cosine similarity difference (IFCS) between features in $E$ and $\tilde{E}$. The WD 
$p$ between two probability distributions $\mathbb{U}$ and $\mathbb{V}$ is:
\begin{equation*}
   W_p(\mathbb{U},\mathbb{V}) = \left( \int_{-\infty}^{+\infty} |\mathbb{U}-\mathbb{V}|^p \right)^{1/p} 
\end{equation*}
where, p is a positive parameter; p = 1 gives the Wasserstein-1 distance (used in our case) Here, `,' in $W_p(\mathbb{U},\mathbb{V})$ denotes that $\mathbb{U}$ and $\mathbb{V}$ are the two probability distributions being compared using the WD. KL between $\mathbb{U}$ and $\mathbb{V}$ is given as:
\begin{equation*}
    D_{KL}(\mathbb{U}||\mathbb{V})=\sum_i (\mathbb{U}(i) \times \log \frac{\mathbb{U}(i)}{\mathbb{V}(i)}
\end{equation*}

Here, `$||$' in $ D_{KL} (\mathbb{U}||\mathbb{V})$ indicate the KL-divergence from $\mathbb{U}$ to $\mathbb{V}$ i.e., a directed measure of divergence from one distribution to another.

Calculating IFCS aims to quantify the loss of information resulting from the compression process. For example, in this case cosine similarity of a feature $f_i$ with all other features $f_j$ in $E$ forms a vector $\mathbb{U}$ and cosine similarity of a feature $f_i$ with all other features $f_j$ in $\tilde{E}$ forms vector $\mathbb{V}$. The features for which the loss is more than a chosen threshold, we use triplet-loss to bring embedding of similar features together. The underlying objective is to train the embedding representation $\tilde{E}$ in such a manner that embeddings of contextually similar features within $E$ are proximally positioned, while embeddings of dissimilar features are distanced from each other. To create triplets (\textit{anchor, positive, negative}) $T_1 , T_2, ..., T_t$ we utilized two techniques: 
\begin{itemize}
    \item \textit{Hard mining}: In this approach, we took a feature for which difference is high, represented as anchor  \( \vec{x}_a \). Positive  \( \vec{x}_p \) is feature that has highest cosine similarity with anchor and negative \( \vec{x}_n \) is the feature that has least cosine similarity. \( \vec{x}_a \), \( \vec{x}_p \) and \( \vec{x}_n \) are selected from the feature set \( X \)=\{\( X_C \), \( X_\mathcal{N} \)\}. Mathematically, this can be expressed as:
\begin{equation*}
\begin{split}
\vec{x}_p = \arg\max{\vec{x}_i} \cos(\vec{x}_a, \vec{x}_i), \\ 
\vec{x}_n = \arg\min{\vec{x}_i} \cos(\vec{x}_a, \vec{x}_i)
\end{split}
\end{equation*}

    \item \textit{Soft mining}: Here, anchor is chosen as in hard mining. Positive is chosen randomly from the subset of features in $E$ having similarity greater than 60\% and having similarity less than 40\% in $\tilde{E}$. Negative is chosen randomly from the subset of features in $E$ having similarity less than 40\% and having similarity greater than 60\% in $\tilde{E}$. 

Formally, let \( E \) and \( \tilde{E} \) be divided into subsets based on cosine similarity with the anchor feature \( \vec{x}_a \).
   \begin{align*}
   u_{E} &= \left\{ \vec{x}_i \in E \mid \cos(\vec{x}_a, \vec{x}_i) > 0.6 \right\} \\
   v_{E} &= \left\{ \vec{x}_i \in E \mid \cos(\vec{x}_a, \vec{x}_i) < 0.4 \right\}\\
   u_{\tilde{E}} &= \left\{ \vec{x}_i \in \tilde{E} \mid \cos(\vec{x}_a, \vec{x}_i) < 0.4 \right\} \\
   v_{\tilde{E}} &= \left\{ \vec{x}_i \in \tilde{E} \mid \cos(\vec{x}_a, \vec{x}_i) > 0.6 \right\}
   \end{align*}
Then, 
   \begin{align*}
  \vec{x}_p \overset{\text{rand}}{\in} \left\{ u_{E} \cup u_{\tilde{E}} \right\} \\
   \vec{x}_n \overset{\text{rand}}{\in} \left\{ v_{E} \cup v_{\tilde{E}} \right\}
   \end{align*}
\end{itemize}
The triplet-loss is computed as $\mathcal{L} = \max(\cos(\vec{x}_a,\vec{x}_p) - \cos(\vec{x}_a,\vec{x}_n) + \alpha,0)$. Here, $\cos(\vec{x}_a,\vec{x}_p)$ represents the cosine similarity between anchor and positive embedding, $\cos(\vec{x}_a,\vec{x}_n)$ represents the cosine similarity between anchor and negative embedding and the margin value $\alpha$ enforces a minimum separation between the positive and negative embedding in the embedding space, ensuring that dissimilar embeddings are adequately distinguished. 

We used two approaches for training MLP: i) replace embedding only for features chosen for triplet mining with the predicted embeddings from model trained on triplets using triplet-loss, ii) replace embedding for all features received after predicting through model trained on triplets and using triplet-loss. Triplets in both approaches are same (for which loss incurred is more as measured using WD and KL). Finally, we used MLP to classify for the presence/absence of PII. 

\subsubsection{PII classification}
\label{PII_classification}
We next infer what type of PII is present in a network flow. PII type classification is a multi-label problem wherein a packet labelled as 1 can contain either one or multiple PIIs flowing through it. We used two approaches: (i) \textit{Leak classification} - to assess how well we infer the PII type from packets that already contain a PII, ignoring packets without PII and (ii) \textit{Combined classification} - assess how well we identify the PII type and the No Leak label, considering all packets. We analyze both datasets which has 23 and 16 PII types respectively.
The emebeddings generated in \ref{PII_detection} and which has best performance for PII detection (in terms of validation/testing accuracy) are passed to MLP to finally get prediction $\{\hat{y}_1,\hat{y}_2, .... \hat{y}_{23}\}$ (in case of ReCon). Next, we explain all the experiments done and results obtained.

\section{Experiments and Results}
\label{Section5}
\subsection{Dataset and preprocessing}
\label{Section3}
We used dataset from two communities ReCon \cite{recon} and AntShield \cite{antsh} summarized in Table \ref{Tab1}.
\subsubsection{ReCon dataset} 
ReCon conducted controlled experiments using Android (5.1.1), iPhone (iOS 8.4.1), and Windows Phone (8.10.14226.359) devices. Each experiment began with a factory reset, followed by connecting the device to Meddle \cite{medd}, which redirected all traffic to a proxy server via VPN tunnels. At the proxy server, software middleboxes intercepted and modified traffic. SSLsplit \cite{sslspl} was used to decrypt and inspect SSL flows during controlled experiments without intercepting human subject traffic.

\subsubsection{AntShield dataset}
This work has collected all packets using AntMonitor \cite{antmo} which is an open-source tool for collecting network traffic from mobile applications. They converted each packet into JSON format and then dissected it into relevant fields such as name  application, server it is contacting, protocol, destination IP and port, headers, payload, timestamp, etc. We observed significantly fewer samples with sensitive data compared to benign samples in this dataset.
\begin{table}
\caption{Summary of used datasets used in our experiments}
\label{Tab1}
\centering
\scriptsize
\begin{tabular}{c|c|c|ccccc} 
\\\hline
\textbf{DataSet} & \textbf{Total} & \textbf{Selected} & \textbf{\#packets} & \textbf{\#leaks} & \textbf{\#non-leaks}  \\
\hline
ReCon &  \multirow{2}{*}{1,428}  &  \multirow{2}{*}{190} & 25,373 & 4,207 & 21,166 \\ \cline{1-1} \cline{4-6}
ReCon-bal\footnotemark[1] & & & 19,683 & 10,033 & 9,650\\\hline
AntShield &  \multirow{2}{*}{554} &  \multirow{2}{*}{171} &29,725 & 7,729 & 21,996  \\ \cline{1-1} \cline{4-6}
AntShield-bal\footnotemark[1] & & & 26,968 & 11,379 & 15,589 \\\hline
\end{tabular}\\
\footnotemark[1]{\textit{bal} represents dataset after applying class balancing algorithm \ref{Algo2}.}\\
\vspace{-0.5cm}
\end{table}
\subsection{Pre-processing}
Let $\mathbb{A}$ be the set of all applications/domains and for each domain $a \in \mathbb{A}$. Inspired by ReCon, the heuristics for selecting domains can be described in Algorithm \ref{Algo1}.

\begin{algorithm}
\footnotesize
\DontPrintSemicolon
\SetAlgoLined
\SetKwInOut{Input}{Input}
\SetKwInOut{Output}{Output}
\Input{$\mathbb{A}$: set of all applications/domains. $thres$: (Optional) Desired number for word count $wc$ (selected empirically as 5).}
\Output{$A\_selected$: set of selected applications.}
\begin{algorithmic}[1]
\FOR{$a \in \mathbb{A}$}
\STATE $n_0 = |a_{label_0}|$ and $n_1 = |a_{label_1}|$ \hfill \emph{// \footnotesize number of non-leak and leak samples}
\STATE $T = n_0+n_0$
\IF{$n_1 >0$ and $n_0 > 0$} 
    \IF{$T \geq 2$ and $n_1 >1$}
        \IF{$wc > thres$}
           \STATE $A\_selected \gets a$
        \ENDIF
    \ENDIF
\ENDIF
\ENDFOR
\end{algorithmic}
\caption{Domain Selection}
\label{Algo1}
\end{algorithm}

\begin{algorithm}
\footnotesize
\DontPrintSemicolon
\SetAlgoLined
\SetKwInOut{Input}{Input}
\SetKwInOut{Output}{Output}
\Input{$D$:  Data-frame containing data with a $label$ column indicating class labels. $F$: (Optional) Desired number of folds 
 for balanced data (default: 10). $M$: (Optional) Threshold for aggressive balancing (default: 5000).}
\Output{$D\_bal$: New data-frame with balanced class distribution.}
\begin{algorithmic}[1]
\STATE \texttt{/*Calculate class imbalance.*/} 
\STATE $n_0 = |D_{label_0}|$ and $n_1 = |D_{label_1}|$ \hfill \emph{// \footnotesize number of class 0  and 1 samples}
\IF{$n_0 < F$ and $n_1 < F$} 
    \STATE  \texttt{/* both classes are too small. */}
    \STATE $\Delta n_0 = F - n_0$ and $\Delta n_1 = F - n_1$ 
\ELSIF{$n_1 > n_0$}
    \STATE  \texttt{/* more positive classes.*/}
    \STATE $\Delta n_0 = n_1 - n_0$
\ELSE 
    \STATE  \texttt{/* more negative classes.*/}
    \IF{$n_0 > M$}
        \STATE \texttt{/*significantly high*/}
        \STATE $\Delta n_0 = M - n_0$
        \ELSIF{$n_0>100$}
        \STATE \texttt{/*moderately high*/}
            \IF{$n_1 > 100$} 
                \STATE $\Delta n_0 = 100 - n_0$.
                \ELSE 
                    \STATE $\Delta n_1 = 100 - n_1$ and $\Delta n_0 = 100 - n_0$
            \ENDIF
        \ELSE
            \STATE $\Delta n_1 = n_0 - n_1$.
    \ENDIF
\ENDIF
\FOR{$i \in \{0,1\}$}
\IF{$\Delta n_i>0$}
\STATE $D_{label_i}^* \gets OverSample(D_{label_i},\Delta n_i)$
\STATE $D_{label_i} \gets Concatenate(D_{label_i}, D_{label_i}^*) $
\ELSIF{$\Delta n_i<0$} 
\STATE $temp = n_0 + n_1 +\Delta n_i$
\STATE $D_{label_i} \gets UnderSample(D_{label_i}, temp)$
\ENDIF
\ENDFOR
\STATE $D\_bal \gets Concatenate(D_{label_1}, D_{label_0})$
\end{algorithmic}
\caption{Class Imbalance Balancing}
\label{Algo2}
\end{algorithm}
The domains selected using algorithm \ref{Algo1} have huge class imbalance as shown in Table \ref{Tab1}. So, we needed to oversample the packets containing sensitive information (leak class). Models constructed using imbalanced data tend to exhibit bias towards predicting observations as members of the majority class. This is due to the model's inclination to give priority to the majority class, which can result in a misleadingly high accuracy. So, inspired by ReCon's method, we used Algorithm \ref{Algo2} to oversample the PII instances and undersample non-PII instances. 
\begin{table*}
    \centering
    \begin{tabular}{>{\centering\arraybackslash}p{0.32\linewidth} >{\centering\arraybackslash}p{0.32\linewidth} >{\centering\arraybackslash}p{0.32\linewidth}}
        \includegraphics[width = 57mm,height=30mm]{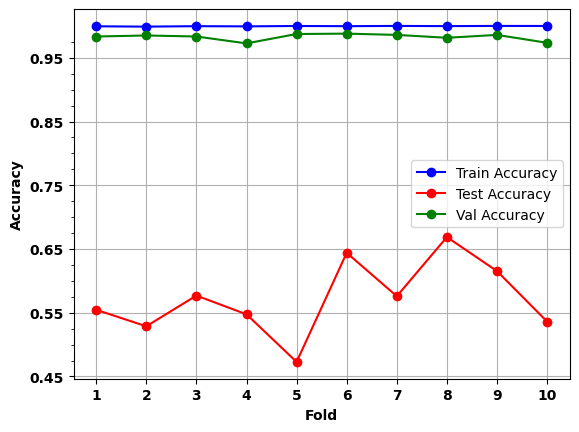} &  \includegraphics[width = 57mm,height=30mm]{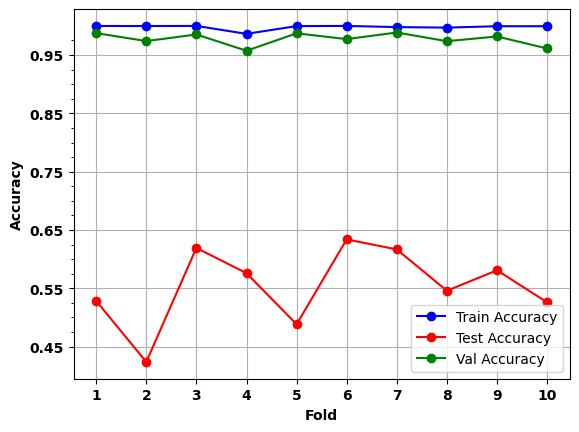} &  \includegraphics[width = 57mm,height=30mm]{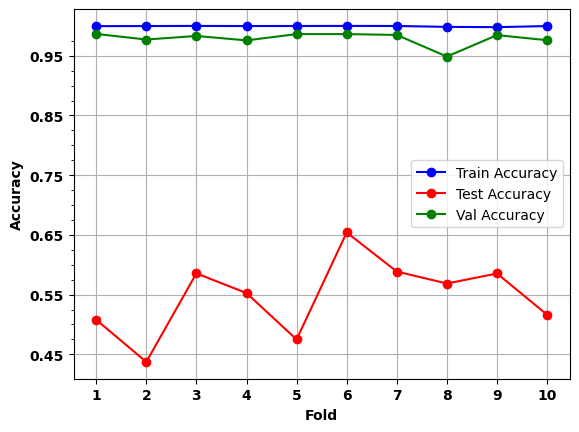} \\
       (a) Evaluation of FT-transformer& 
        (b) reducing number of neurons& 
        (c) using L1 regularization\\
        \includegraphics[width = 57mm,height=30mm]{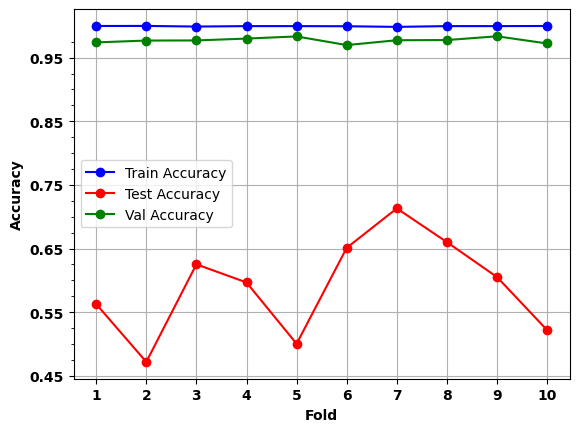} &  \includegraphics[width = 57mm,height=30mm]{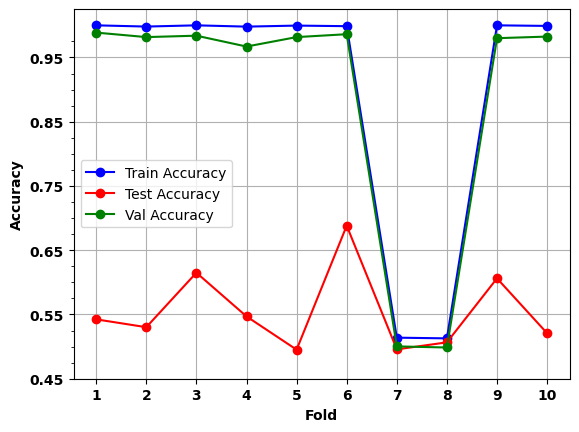} &  \includegraphics[width = 57mm,height=30mm]{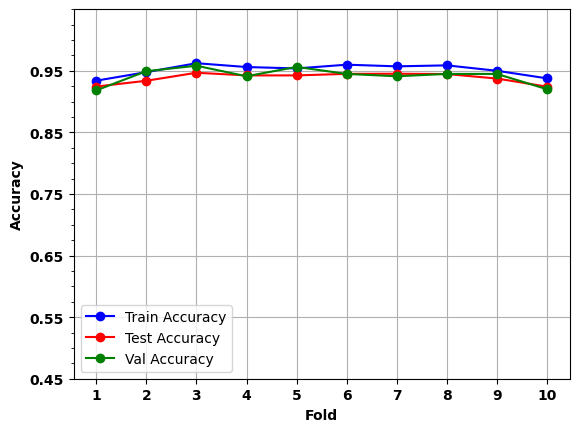} \\
        (d) using L2 regularization & 
        (e) using drop-outs & 
        (f) evaluation of End-to-End learning framework (after PCA)\\
    \end{tabular}
     \captionof{figure}{10-fold cross validation plots for ReCon dataset}
    \label{fig-2}
\end{table*}
\subsection{Results}
The experiments were initially conducted on the ReCon dataset and subsequently AntShield. When converted to a tabular format, ReCon contains 19,683 samples, when balanced and after eliminating any features that have a single value across all samples (if present), dataset has  692 features. Similarly, the AntShield has 26,968 samples and 743 features. If a feature value is missing in any sample, it is replaced with `-' for categorical features and with `0' for numerical features. This replacement is necessary because the FT-Transformer does not function with null values. The dataset $D$ here is divided into disjoint sets for $D_{\text{train}}$ (used for model training), $D_{\text{val}}$ (used for validation i.e., hyper-parameter tuning and early stopping), and $D_{\text{test}}$ (used for final model evaluation). k-fold cross-validation is used for model evaluation. In it, the data $D$ is split into $k$ equally sized folds. For each of the $k$ iterations, one fold is used for validation while the remaining $k-1$ folds are used for training. After $k$ iterations, all data points have been used for both training and validation. $D_{\text{test}}$ remains separate and is used for final model evaluation after the cross-validation process. We assume that all sets are mutually exclusive, meaning no data point belongs to more than one set at a time.

\subsection{Using FT-Transformer}
The embeddings $E$ of dimension $Q=32$ together with $CLS$ token are inputted to one transformer layer with eight attention heads (chosen empirically). Transformer converts $E_{FT}$ to contextualized embedding $E'_{FT}$ which is then inputted to MLP with first layer as normalization. The second layer is a dense layer with neurons=$\left\lfloor \frac{\text{Q}}{2} \right\rfloor$ and activation as ReLU. The output layer has sigmoid activation as the problem is a binary classification problem. The loss function used here is binary-cross entropy (BCE). The testing accuracy in our case came out to be higher than training accuracy. This is because training set is more diverse or contains more challenging samples, while the test set is more representative of simpler cases. So, we also evaluated our framework with 10-fold cross validation so as to check that model's performance is consistent and stable across different data splits. Results in Figure \ref{fig-2}(a) show that the framework is over-fitted with 99.96\% training, 98.26\% validation and 57.21\% testing accuracy in case of ReCon dataset. So to reduce the over-fitting, we tried different techniques: 
\subsubsection{Reduce model complexity} We reduced model's complexity by decreasing the number of neurons in second layer to $Q//4$. This gave 99.80\% training and 55.38\% testing accuracy. The model still over-fitted.
\subsubsection{Using regularization constraints}
L1 regularization and L2 regularization are two popular techniques used to mitigate overfitting in a model. In the case of L1 regularization, the loss function is formulated as: $\mathcal{L}_{new}=\mathcal{L}_{BCE}+\lambda \sum_{i=1}^n |\theta_i|$. For L2 regularization, the loss function is given by: $\mathcal{L}_{new}=\mathcal{L}_{BCE}+\lambda \sum_{i=1}^n \theta_i^2$. Here, $\mathcal{L}_{BCE}$ represents the binary cross-entropy loss (loss used in \cite{fttrans}), $\lambda$ is the regularization parameter that controls the strength of the regularization, and $\theta$ denotes the model’s weights. $\lambda$, here in our experiments is $0.01$. It is seen in case of L1, the training and test accuracies are 99.94\% and 54.71\% , whereas using L2, the training and test accuracies are 99.95\% and 59.08\% respectively. Again, model is over-fitted.
\subsubsection{Using dropouts} Regularization methods such as L1 and L2 mitigate overfitting by altering the cost function. In contrast, the dropout technique modifies the network architecture itself to prevent overfitting. During training, dropout randomly deactivates a subset of neurons (excluding the output layer) in each iteration. The probability of each neuron being dropped is determined by a predefined dropout rate. For instance, we choose dropout rate $p=0.25$, therefore probability $\mathbb{P}$(neuron dropped)$=0.25$. During training, for each neuron $i$, we have: 
\begin{table}
    \centering
    \scriptsize
    \caption{Summary of binary classification results for FT-transformer.}
    \begin{minipage}{\linewidth} 
    \centering
    \begin{tabular} {|p{0.85cm}|p{0.5cm}|p{0.5cm}|p{0.5cm}|p{0.5cm}|p{0.5cm}|p{0.5cm}|p{0.5cm}|} \hline
     \multirow{2}{*}{\textbf{DataSet}}  & \multirow{2}{*}{\textbf{Crit.\footnotemark[1]}} &  \multicolumn{3}{c|}{\textbf{without K-fold}} &  \multicolumn{3}{c|}{\textbf{with K-fold}} \\ \cline{3-8}
    & & \textbf{Train} & \textbf{Valid.} & \textbf{Test} & \textbf{Train} & \textbf{Valid.} & \textbf{Test} \\ \hline
    \multirow{5}{*}{ReCon} & - & 50.71 & 98.38 & 97.71 & 99.96 & 98.26 & 57.21 \\ \cline{2-8}
    & RC\footnotemark[2] & 49.96 & 98.22& 97.56 & 99.80& 97.75 & 55.38\\ \cline{2-8}
    & L1 & 50.41& 97.24& 96.52& 99.94 & 97.89 & 54.71 \\\cline{2-8}
    & L2 & 49.40& 98.25& 97.74& 99.95 & 97.72 &59.08\\ \cline{2-8}
    & Drop.\footnotemark[3] & 50.15 & 97.81&97.64 & 90.20 & 88.51 & 55.47\\ \hline
    \multirow{5}{*}{AntShield} & - & 51.79 & 98.52 & 98.54 & 99.93 & 99.17& 65.23\\ \cline{2-8}
    & RC\footnotemark[2] & 51.44 &99.05 & 98.89& 99.90 & 99.10& 65.12\\ \cline{2-8}
    & L1 & 51.22 &99 & 98.87& 99.91 & 99.84& 62.42\\ \cline{2-8}
    & L2 & 51.34 &99.05 & 98.78& 99.92 & 99.11& 64.63\\ \cline{2-8}
    & Drop.\footnotemark[3] & 51.32 &98.7 & 98.65& 99.91 & 99.02& 62.22\\ \hline
    \end{tabular}
    \footnotemark[1]{Criteria};
      \footnotemark[2]{Reduced complexity};
       \footnotemark[3]{Drop-outs}.\\
    \end{minipage}
    \label{Tab-1}
\end{table}
\begin{equation*}
    \tilde{h}_i^{(l)} = \begin{cases}0 & \text{with\ probability $p$}\\ 
    \frac{h_i^{(l)}}{1-p} & \text{with probability $1-p$}\end{cases}
\end{equation*} 
where $ \tilde{h}_i^{(l)}$ is the output of neuron $i$ in layer $l$ after applying dropout, and $h_i^{(l)}$ 
is the original output of neuron $i$ in layer $l$ before dropout. This scaling by $\frac{1}{1-p}$ ensures that the expected sum of the outputs of the neurons remains the same during training as it would without dropout. This adjustment helps to maintain the overall output magnitude consistent between training and inference. Using drop-outs, the training and test accuracies are 90.20\% and 55.47\% respectively. It is seen that model's over-fitting is not reduced.
\begin{table}
    \centering
    \begin{tabular}{>{\centering\arraybackslash}p{0.47\linewidth} >{\centering\arraybackslash}p{0.47\linewidth}}
        \includegraphics[width=42mm, height=25mm]{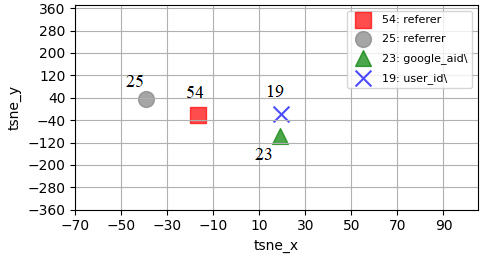} &  \includegraphics[width=42mm, height=25mm]{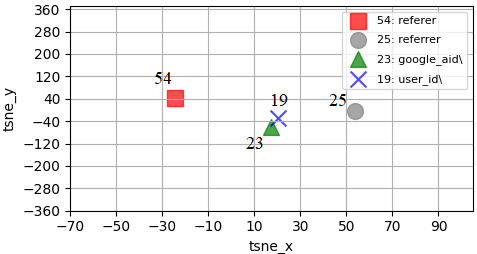} \\
        (a) before passing to transformer & 
        (b) after passing to transformer\\
        \includegraphics[width=42mm, height=25mm]{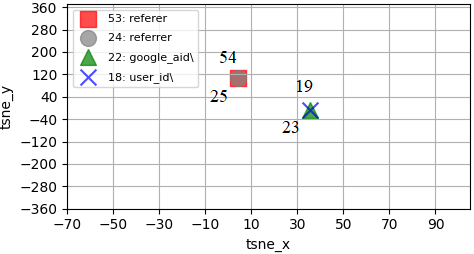} &  \includegraphics[width=42mm, height=25mm]{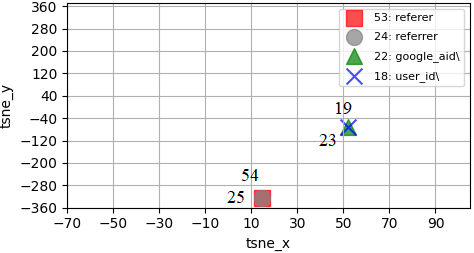} \\
        (c) after using end-to-end learning & 
        (d) using WD, HM-1\\
        \includegraphics[width=42mm, height=25mm]{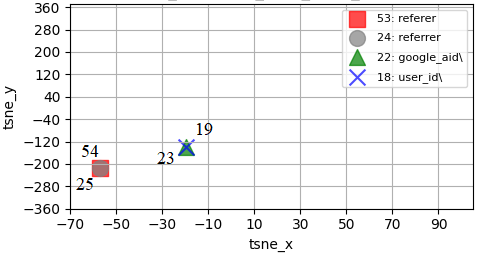} &  \includegraphics[width=42mm, height=25mm]{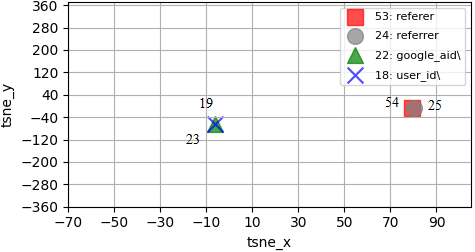} \\
        (e) using WD, HM-2 & 
        (f) using WD, SM-1\\
        \includegraphics[width=42mm, height=25mm]{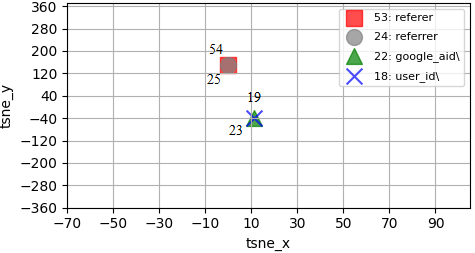} &  \includegraphics[width=42mm, height=25mm]{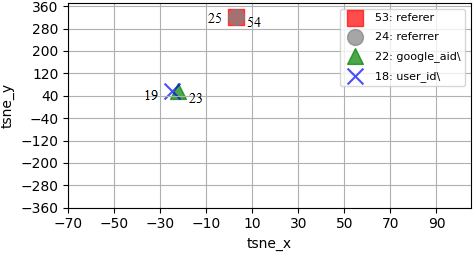} \\
        (g) using WD, SM-2 & 
        (h) using KL, HM-1\\
        \includegraphics[width=42mm, height=25mm]{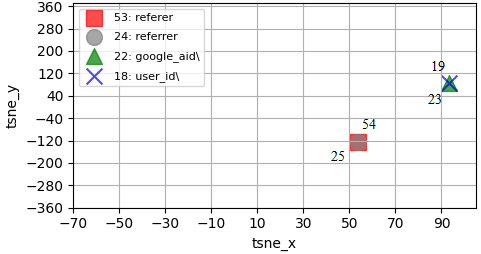} &  \includegraphics[width=42mm, height=25mm]{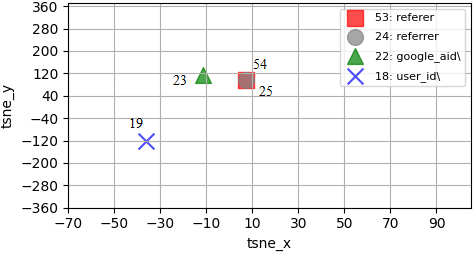} \\
        (i) using KL, HM-2 & 
        (j) using KL, SM-1\\
        \multicolumn{2}{c}{
            \includegraphics[width=42mm, height=25mm]{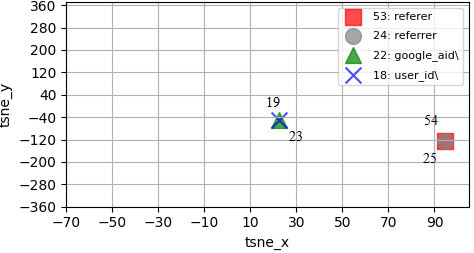}
        } \\
        \multicolumn{2}{c}{
            (k) using KL, SM-2.
        } 
    \end{tabular}
    \captionof{figure}{For ReCon dataset: t-SNE plots for 4 features: `referer',  `referrer' and `google\_aid', `user\_id'. Abbreviations used are: HM: Hard-mining, SM: Soft-mining, 1: Replaced embedding of features selected for triplets, 2: Replaced all features with predicted embeddings.}
\vspace{-0.9cm}
    \label{fig-6}
\end{table}

Therefore, all these approaches failed to reduce over-fitting for ReCon dataset as shown in Figure \ref{fig-2} (b, c, d and e). All the results for ReCon and AntShield dataset have been summarized in Table \ref{Tab-1}. Intuitively, we analyzed the embedding using t-SNE plots as shown in Figure \ref{fig-6} (a and b). The results show that the transformer here failed to bring embedding for related features closer. For instance, features such as `referer' and `referrer' is HTTP header field that identifies the address of the web page from which the resource has been requested. Second case considered here is, `google\_aid' and `user\_id' which refer to the advertising id used as device identifier for advertisers that allows them to measure user ad activity on user's devices. Embeddings for similar features within these two groups are far from each other i.e., the embeddings for `referer' and `referrer' are distant, as are those for `google\_aid' and `user\_id'. To identify the issue, we examined the code given in \cite{fttrans}, and found that the authors utilize the Keras Embedding class, keras.layers.Embedding, repetitively for each feature. Consequently, this results in the initialization of random weights for each feature, irrespective of their semantic similarities. So, in-spite of having same value, embeddings for features in these two groups have distinct embeddings, failing to capture their semantic relationship as shown in Figure \ref{fig-6}(a). Consequently, when these embeddings are passed to the transformer model, it impairs the model's ability to infer meaningful contextual relationships between such features as depicted in Figure \ref{fig-6}(b). Next, we explain the results obtained from end-to-end learning approach.
\begin{figure*}
    \centering
\includegraphics[width=0.9\linewidth, height=0.35\linewidth]{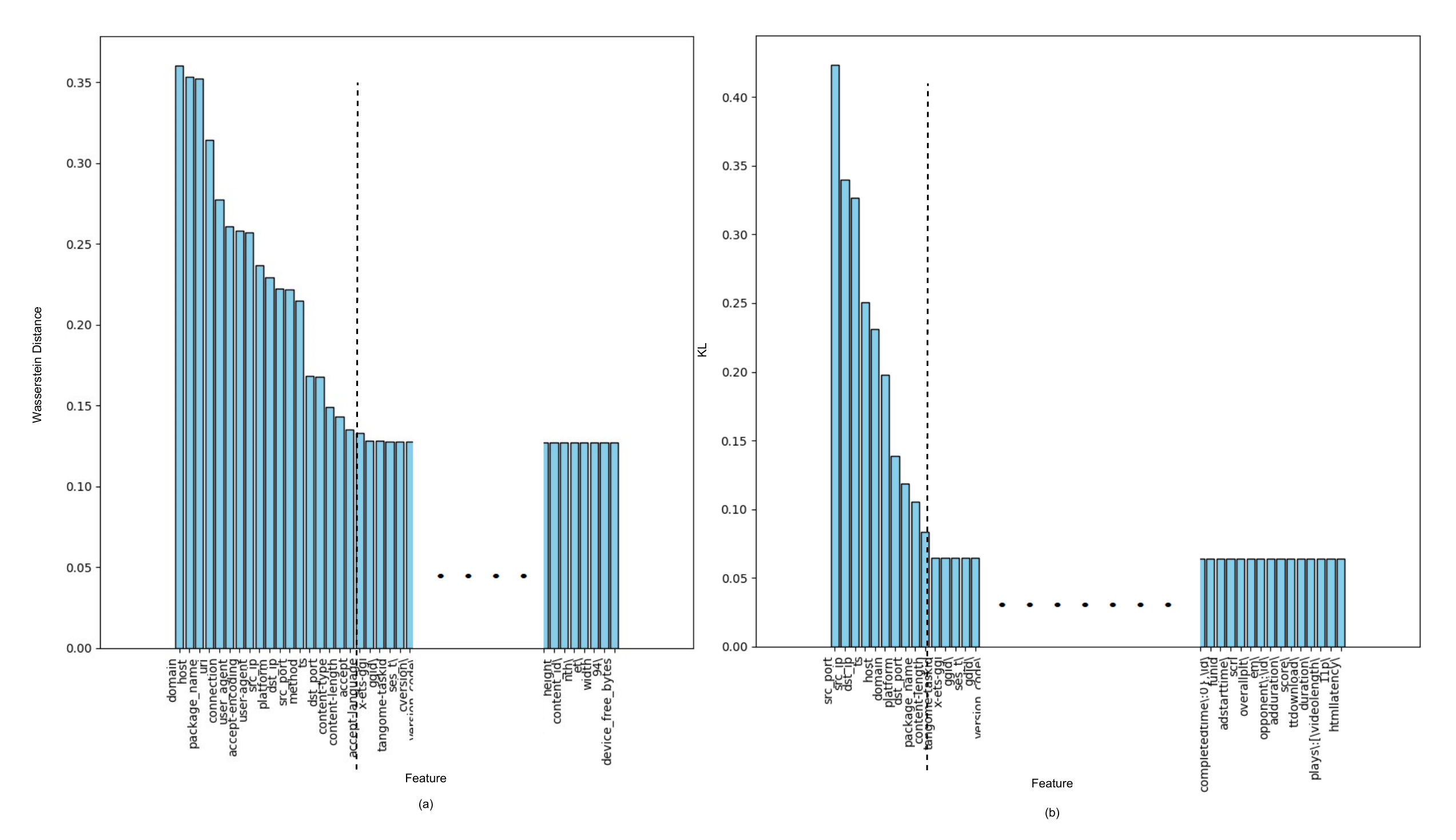}
    \caption{Measurement of distributional difference between $E$ and $\tilde{E}$ using (a) WD (b) KL for ReCon data-set. Features left to the  dotted line are chosen for triplet mining.}
    \label{fig-4}
\end{figure*}

\begin{table*}
    \centering
    \caption{Summary of binary classification results for End-to-End learning.}
    \begin{minipage}{\linewidth}\centering
    \scriptsize 
    \begin{tabular}{|p{1cm}|p{0.55cm}|p{0.45cm}|p{0.55cm}|p{0.55cm}|p{0.55cm}|p{0.55cm}|p{0.55cm}|p{0.55cm}|p{0.55cm}|p{0.45cm}|p{0.55cm}|p{0.55cm}|p{0.55cm}|p{0.55cm}|p{0.55cm}|p{0.55cm}|}
        \hline
        \multirow{3}{*}{\textbf{DataSet}}&\multirow{3}{*}{\textbf{Dist.\footnotemark[1]}} & \multirow{3}{*}{\textbf{n\_T\footnotemark[2]}} & \multirow{3}{*}{\textbf{Crit.\footnotemark[3]}} & \multicolumn{6}{c|}{\textbf{without PCA}} & \multicolumn{7}{c|}{\textbf{with PCA}} \\ \cline{5-17}
           & & & & \multicolumn{3}{c|}{\textbf{without K-fold}} & \multicolumn{3}{c|}{\textbf{with K-fold}} &   \multirow{2}{*}{\textbf{n\_C\footnotemark[4]}} & \multicolumn{3}{c|}{\textbf{without K-fold}} & \multicolumn{3}{c|}{\textbf{with K-fold}} \\ 
            \cline{5-10} \cline{12-17}
            & & & & \textbf{Train} & \textbf{Valid.} & \textbf{Test} & \textbf{Train} & \textbf{Valid.} &\textbf{Test} & & \textbf{Train} & \textbf{Valid.} & \textbf{Test} & \textbf{Train} & \textbf{Valid.} & \textbf{Test} \\ \hline
        
        \multirow{9}{*}{ReCon} & -&-&- & 82.62&83.08&81.92&81.02&80.45&79.38& 280&93.47&92.41&92.15&95.16&94.18&93.86\\ \cline{2-17}
        
         & \multirow{4}{*}{WD} & \multirow{4}{*}{\centering 19} & HM\footnotemark[5] & 65.87&65.30&63.66&70.38&70.36&69.06& 249&83.01&82.44&80.65&85.96&85.23&87.00 \\ 
         
         & & & HM\footnotemark[6] & 65.36&64.86&64.53&62.90&62.18&62.04&259&89.44&87.05&86.85&92.53&90.89&90.86 \\ 
         
         & &  & SM\footnotemark[5] & 71.66&70.98&71.10&70.14&70.43&68.72&	286&90.09&88.92&88.45&93.34&92.24&93.45\\ 
         
         & & & SM\footnotemark[6] & 52.18&51.62&53.17&68.56&68.77&66.86&	234&89.22&87.46&87.71&92.97&91.59&92.94\\ \cline{2-17}
         
        & \multirow{4}{*}{KL} & \multirow{4}{*}{\centering 10} & HM\footnotemark[5] & 76.44&76.92&74.66&80.35&79.77&79.05&280&91.82&90.57&89.99&95.52&94.09&94.31 \\ 
        
        & &  &HM\footnotemark[6] & 68.04&67.40&67.85&68.94&69.19&67.97&338&\textbf{94.56}&\textbf{93.17}&\textbf{93.04}&\textbf{96.57}&\textbf{95.12}&\textbf{95.68}\\ 
        
        & & & SM\footnotemark[5] & 78.30&77.43&77.32&78.82&78.62&77.31&202&84.86&85.02&83.37&86.31&85.39&86.64\\ 
        
        & &  & SM\footnotemark[6]& 69.40&69.40&67.29&67.60&67.63&65.63&123&72.72&72.10&71.18&73.13&72.69&72.12\\ \hline
        
        \multirow{9}{*}{AntShield} & -&-&- & 91.05&90.78&90.66&90.53&90.49&90.15& 320&94.03&93.42&93.05&95.12&94.36&94.83  \\ \cline{2-17}
        
         & \multirow{4}{*}{WD} & \multirow{4}{*}{\centering 18} & HM\footnotemark[5] & 81.21&80.56&80.18&85.29&85.35&84.99 & 355&86.93&86.26&86.08&88.22&87.91&87.82  \\ 
         
         & & & HM\footnotemark[6] & 70.54&69.71&70.69&74.58&74.61&73.98& 202&92.11&90.50&90.36&94.67&93.49&94.20  \\ 
         
         & &  & SM\footnotemark[5] & 76.33&77.15&77.03&84.98&84.55&84.33& 311&92.02&91.61&90.79&93.24&92.61&93.05	  \\ 
         
         & & & SM\footnotemark[6] & 75.55&73.86&74.42&81.52&81.35&80.60& 161&90.94&90.01&90.43&93.25&92.58&92.90	 \\ \cline{2-17}
         
        & \multirow{4}{*}{KL} & \multirow{4}{*}{\centering 4} & HM\footnotemark[5] & 41.78&43.36&42.58&64.98&65.01&64.39&319&93.49&93.21&92.64&94.67&93.96&94.10\\ 
        
        & &  &HM\footnotemark[6] &  76.64&76.29&76.10&76.61&76.61&75.55&338&95.05&93.37&94.25&96.40&95.29&95.88 \\ 
        
        & & & SM\footnotemark[5] &73.12&71.94&71.97&73.55&73.55&73.06&222&92.58&91.89&91.16&93.20&92.71&93.14\\ 
        
        & &  & SM\footnotemark[6]&74.29&72.98&73.10&81.43&81.00&80.57&  395&\textbf{95.78}&\textbf{94.86}&\textbf{94.49}&\textbf{97.36}&\textbf{96.16}&\textbf{96.90}  \\ \hline
    \end{tabular} \\
    \begin{minipage}{\textwidth}
        \raggedright
     \footnotemark[1]{Distance used to compute distributional difference.};
      \footnotemark[2]{Number of Triplets};
      \footnotemark[3]{Criteria for evaluation. HM denote hard-mining and SM denote soft-mining};
    \footnotemark[4]{Number of principal components};
        \footnotemark[5]{Replaced embedding of features selected for triplets};
        \footnotemark[6]{Replaced all features with predicted embeddings.}
        \end{minipage}
    \end{minipage}
    \label{Tab-3}
\end{table*}

\begin{figure}
    \centering
    \begin{minipage}[b]{0.48\columnwidth}
    \includegraphics[width=\textwidth,height=0.8\textwidth]{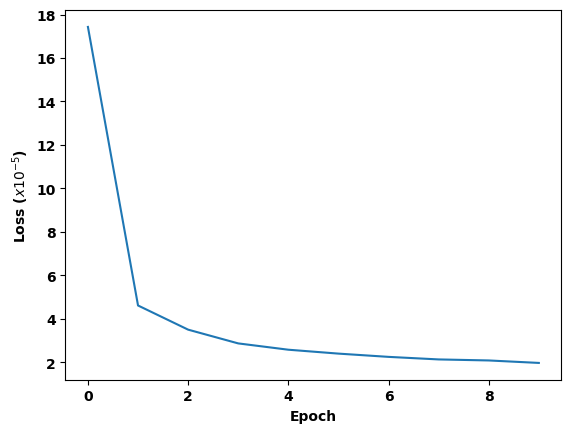}
        \caption*{a) ReCon}
        \label{fig:figure1}
    \end{minipage}
    \hfill
    \begin{minipage}[b]{0.48\columnwidth}
        \includegraphics[width=\textwidth,,height=0.8\textwidth]{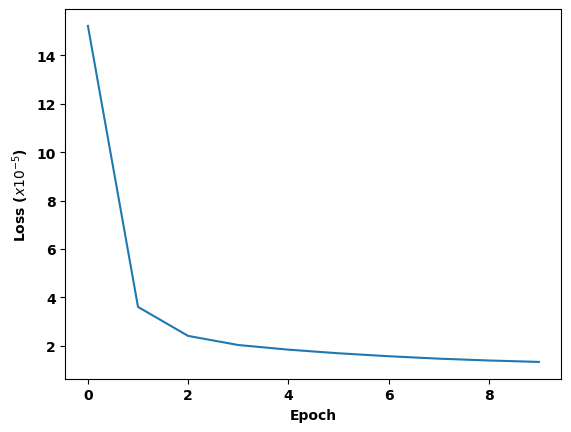}
        \caption*{b) AntShield}
    \end{minipage}
     \label{losses_ae_both}
    \caption{Autoencoder training MSE loss}
\end{figure}

\subsection{Using End-to-End learning Framework}
In this approach, we used pre-trained model provided by \cite{sbert}. Specifically we utilized `all-MiniLM-L6-v2' which is 5 times faster of all trained models in \cite{sbert} and still offers good quality. This model is lightweight but precise, allowing it to provide a 384-dimensional embedding for each feature's value. This model creates vectors that represent each feature's value in a way that captures their semantic relationships. Similar attributes will be closer in this vector space, reflecting their semantic similarity. But processing this high dimensional size of embeddings $E$ generated require huge computation power. Due to lack of computational resources, autoencoder is used to convert these embeddings to a condensed representation $\tilde{E}$ having 32 dimension (as in case of FT-transformer) with minimum reconstruction loss. 
\begin{table}
    \centering
    \begin{tabular}{>{\centering\arraybackslash}p{0.47\linewidth} >{\centering\arraybackslash}p{0.47\linewidth}}
        \multirow{2}{*}{
            \includegraphics[width=42mm, height=25mm]{figs/tsne_plots/corrected_embs_soft_kl_new.png}
        } &  \includegraphics[width=42mm, height=25mm]{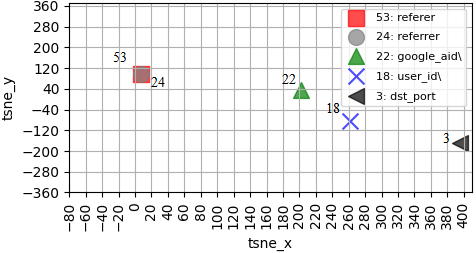} \\
        & (b) \\
        & \includegraphics[width=42mm, height=25mm]{figs/tsne_plots/autoencoder.png} \\
        (a) & (c) \\
    \end{tabular}
    \captionof{figure}{t-SNE plots justifying the second group embeddings `google\_aid', `user\_id' moving apart in Figure \ref{fig-6}-(j) i.e., using KL, soft-mining and replacing embedding of features selected for triplets. (a) showing Figure \ref{fig-6}-(j) embedding, (b) showing embedding of anchor `dst\_port' along with other 4 features in same case and (c) showing $\tilde{E}$ for all 5 features.}
     \vspace{-0.5cm}
    \label{fig-6(reason)}
\end{table}
The autoencoder here consist of encoder with 1 input layer and 2 hidden layers, having 128 and 64 neurons each and rectified linear unit (ReLU) as the activation function. Then is the bottleneck layer having 32  neurons that gives the latent representation of the input embedding. The bottleneck is succeeded with a decoder unit. Decoder has an architecture similar to encoder because we have to reconstruct the input. The reconstruction loss used here is mean squared error. Figure 7 shows the loss encountered for training. This latent representation of embeddings $\tilde{E}$ is flattened to a 2-D representation and then passed to a MLP. 

MLP used has three layers: first is a dense layer with 128 neurons and ReLU activation; second is also a dense layer with 64 neurons and ReLU activation; and the output layer has sigmoid activation as the problem is a binary classification problem. We evaluated our framework with and without 10-fold cross validation as shone in Table \ref{Tab-3}. However, the flattened embeddings suffer from curse of dimensionality as the number of samples $\leq$ number of features. In the case of the ReCon dataset, the flattened embeddings have a shape of (19683, 692*32), where 19683 represents the number of samples, 692 denotes the number of features, and 32 is the dimension of each feature value. Whereas, in case of AntShield dataset, the flattened embeddings have a shape of (26,968, 743*32). 
\begin{table*}
    \centering
    \begin{tabular}{>{\centering\arraybackslash}p{0.32\linewidth} >{\centering\arraybackslash}p{0.32\linewidth} >{\centering\arraybackslash}p{0.32\linewidth}}
        \includegraphics[width = 57mm, height =30mm]{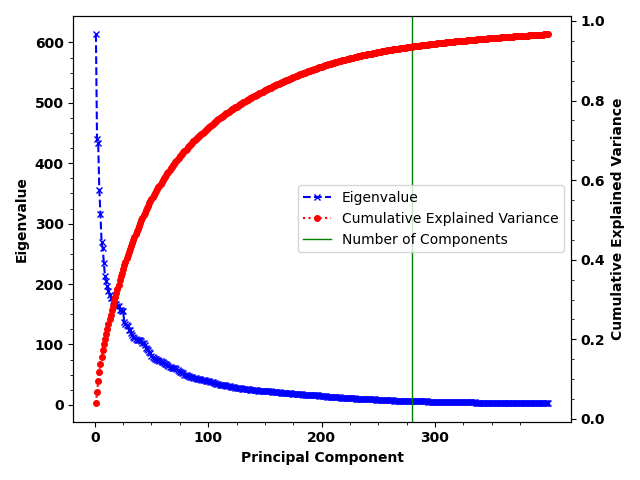} &  \includegraphics[width = 57mm, height =30mm]{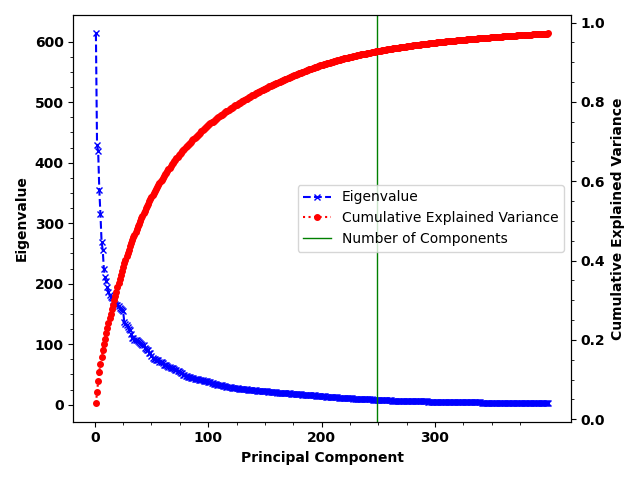} &  \includegraphics[width = 57mm, height =30mm]{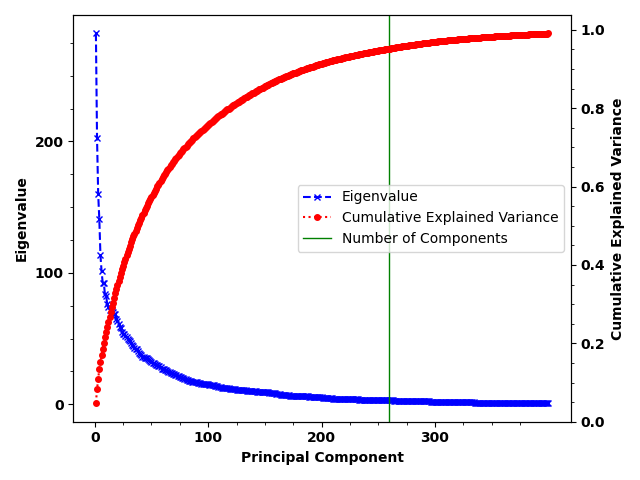} \\
       (a) without TL, n\_C=280& 
        (b) with TL, WD, HM-1, n\_C=249& 
        (c) with TL, WD, HM-2, n\_C= 259\\
        \includegraphics[width = 57mm, height =30mm]{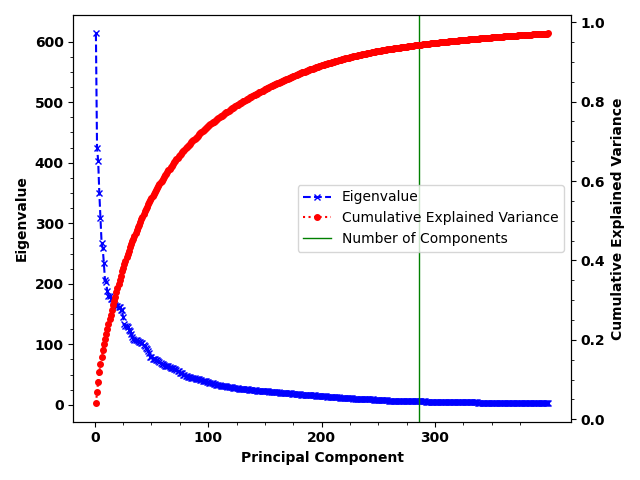} &  \includegraphics[width = 57mm, height =30mm]{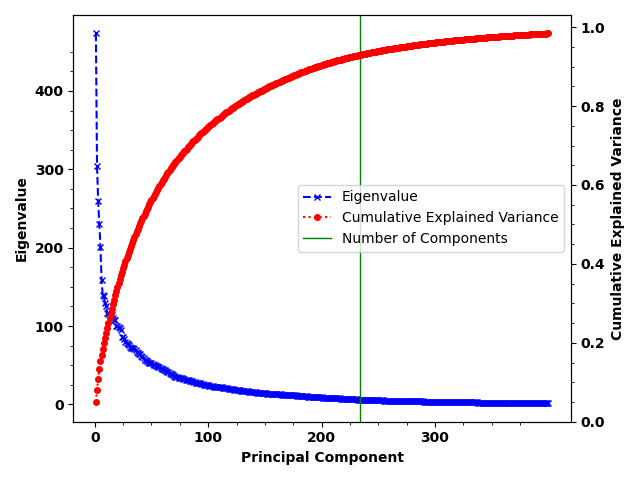} &  \includegraphics[width = 57mm, height =30mm]{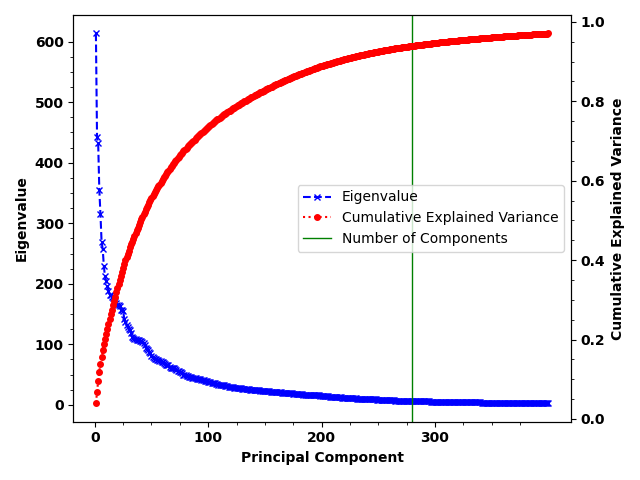} \\
        (d) with TL, WD, SM-1, n\_C=286 & 
        (e) with TL, WD, SM-1, n\_C=234 & 
        (f) with TL, KL, HM-1, n\_C=280\\
        \includegraphics[width = 57mm, height =30mm]{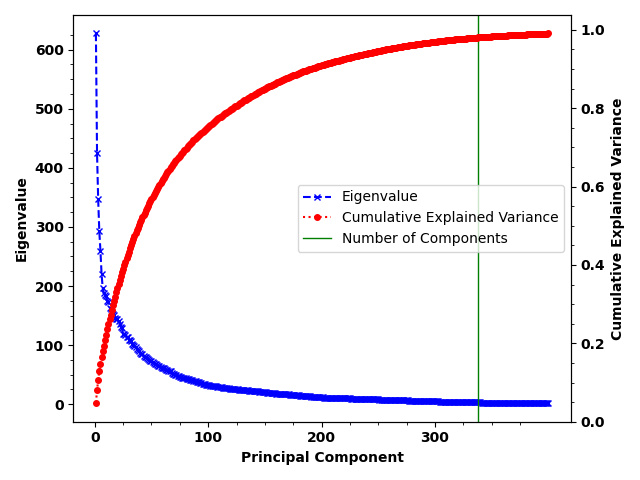} &  \includegraphics[width = 57mm, height =30mm]{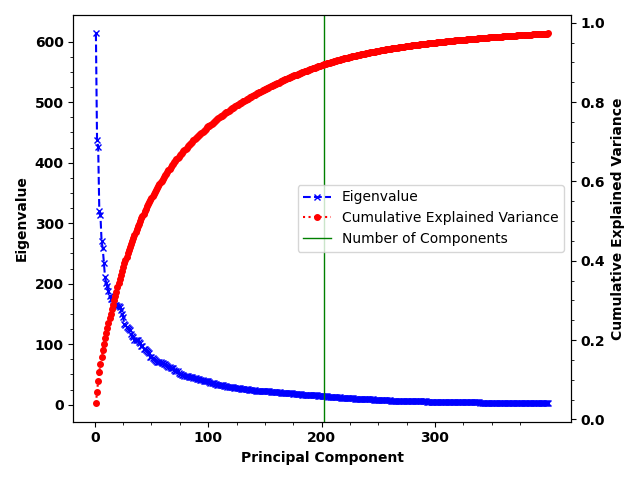} &  \includegraphics[width = 57mm, height =30mm]{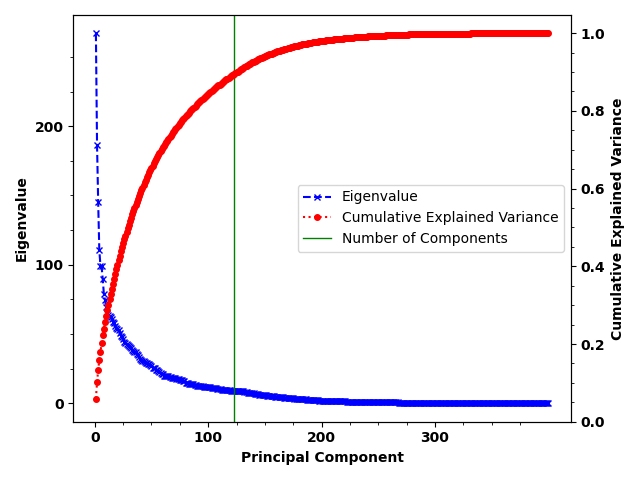}\\
        (g) with TL, KL, HM-2, n\_C=338 & 
        (h) with TL, KL, SM-1, n\_C=202 & 
        (i) with TL, KL, SM-2, n\_C=123 \\
    \end{tabular}
     \captionof{figure}{Scree plots for ReCon dataset embeddings ($\tilde{E}$) after using end-to-end learning to select principal components. Here, abbreviations used are:- TL: Triplet-Loss, HM: Hard-mining, SM: Soft-mining, 1: Replaced embedding of features selected for triplets, 2: Replaced all features with predicted embeddings, n\_C: Number of principal components selected.}
    \label{eigenplots}
\end{table*}

\begin{table*}
    \centering
    \begin{tabular}{>{\centering\arraybackslash}p{0.47\linewidth} >{\centering\arraybackslash}p{0.47\linewidth}}
        \includegraphics[clip, trim=0.18cm 0.2cm 0.18cm 1.1cm,width = 75mm,height=50mm]{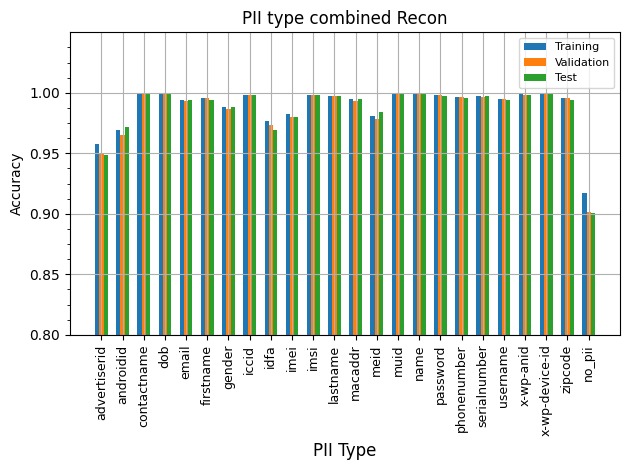} &  
        \includegraphics[clip, trim=0.18cm 0.2cm 0.18cm 1.1cm, width = 75mm,height=50mm]{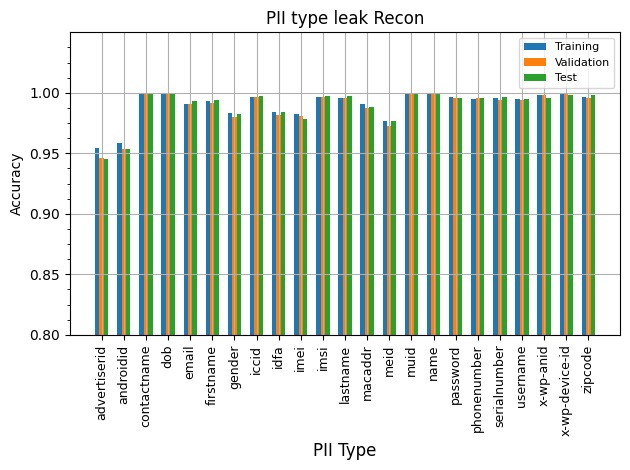} \\
        (a) ReCon: Combined approach & (b) Recon: Leak detection approach\\
        \includegraphics[clip, trim=0.18cm 0.2cm 0.18cm 1.1cm,width = 75mm,height=50mm]{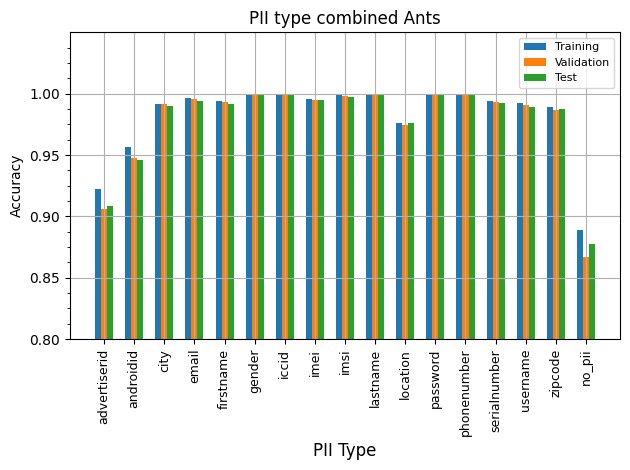} &  
        \includegraphics[clip, trim=0.18cm 0.2cm 0.18cm 1.1cm,width = 75mm,height=50mm]{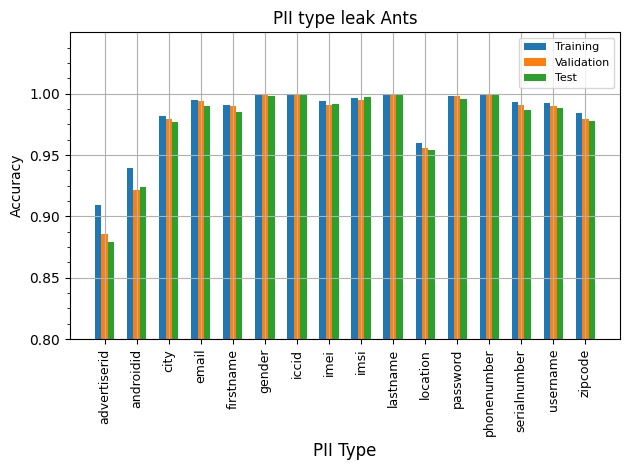} \\
        (c) AntShield: Combined approach & (d) AntShield: Leak detection approach\\
    \end{tabular}
    \captionof{figure}{PII type classification.}
    \label{fig7}
\end{table*}
Therefore, we applied PCA to reduce dimensionality after normalizing the features. We used normalization before applying PCA to the flattened embeddings to ensure that each feature contributes equally to the analysis. Without normalization, features with larger scales could dominate the principal components, leading to biased results. By standardizing the data, we rescaled the features to have a mean of zero and a standard deviation of one, allowing PCA to identify the true directions of maximal variance. Further, to determine the number of principal components, we used scree plot, which shows the amount of variation captured by each component, as illustrated in Figure \ref{eigenplots} (showing first 400 principal components due to space constraints). We identified the principal components that explain the maximum variance in the embeddings by finding the elbow point on the plots using kneedle algorithm \cite{kneedle}. Using these selected principal components, we effectively reduced the dimensionality of the embeddings. The reduced dataset is subsequently fed into the MLP, resulting in improved accuracies of 95.17\%, 94.18\%, and 93.86\% for training, validation, and testing, respectively with 280 principal components for ReCon dataset as shown in Figure \ref{fig-2}(f) and Table \ref{Tab-3}. 

After getting the 32 dimension $\tilde{E}$ from 384 dimension embedding $E$, we quantity the loss that occurred in this conversion using IFCS as discussed in Section \ref{PII_detection} and shown in Figure \ref{fig-4}. Features having high difference (left to dotted line) in are taken for further processing. We employed triplet loss to further process the embedding of features selected above. We used hard and soft mining techniques as discussed in Section \ref{PII_detection} to create triplets. Triplets created are then used to train the model which then predicts the embeddings wherein anchor is positioned near to positive and away from negative. Table \ref{Tab-3} depicts summarized results and Figure \ref{fig-6}(c-k) shows t-SNE plots of embeddings for all scenarios. Hard-mining with KL-Divergence as distance metric to quantify loss gave best results with 96.57\% training and 95.68\% test accuracy in case of k-fold cross validation using PCA with replacing all features predicted from model trained on features selected for triplet-mining using triplet loss. 
As depicted in Figure \ref{fig-6}(j), the embeddings for second group `google\_aid' and `user\_id' have moved apart in case of using KL Divergence as metric to measure the distributional difference between $E$ and $\tilde{E}$, soft-mining as triplet mining approach and replacing embedding of only features selected for triplets and rest same of $\tilde{E}$. We investigated the reason behind this movement (depicted in Figure \ref{fig-6(reason)}) and found that `user\_id' became part of one of the triplet as negative to anchor `dst\_port' (destination port). So, `user\_id' moved far from `dst\_port' and also from `google\_aid' as shown in Figure \ref{fig-6(reason)}(b) compared to $\tilde{E}$ where `dst\_port' was close to both `user\_id' and `google\_aid' Figure \ref{fig-6(reason)}(c).  

\subsection{PII Types}
PII type classification is a multi-label problem wherein each flow can have multiple PII types flowing through it, so we used MultiLabelBinarizer\cite{multilabel}, a utility from the scikit-learn library that converts the list of labels into a binary matrix. This matrix indicates the presence/absence of each type, with a 1 or 0 for each possible type per flow. The embeddings $\tilde{E}'$ are fed into $h_{\tau}$ with the binary matrix as the target. MLP outputs $\{\hat{y}_1,\hat{y}_2, .... \hat{y}_{23}\}$ (in case of ReCon) where $\hat{y}_t$ is binary prediction for each PII type. The model is evaluated based on its ability to infer the PII type from packets that already contain a PII, ignoring packets without PII. It is also assessed on combined classification, which measures how well the model identifies both the PII type and the no leak label `no\_pii', considering all packets. Figure \ref{fig7} shows the model's performance for each PII type in both scenarios on both datasets. The results show average accuracy across 10 folds for each PII type. For some types accuracy is high because certain types are easy to learn and get near 100\%, while a small set of leak types are difficult. One limitation of applied PII type detection approach is that the model's detection capability is restricted to the PII types present in the training data. If a type appears in the test data but not in the training data, the model will fail to detect it. This issue will be addressed with class-incremental learning in our future work.

\section{Conclusion}
\label{Section7}
This paper proposes a novel end-to-end learning framework for mobile packet classification to detect PII exposure and evaluates its efficiency and effectiveness  using two real-world datasets. First we evaluated the performance of a state-of-art framework that works well for tabular dataset. We then proposed a deep learning framework for predicting PII exposure from user devices, employing a triplet-loss based fine-tuning method to enhance detection capability. We have shown that our framework achieves higher accuracy compared to state-of-the-art works on PII detection \cite{recon,antsh, ants, SPCOM}.

\textbf{Future work.} There are many directions for future work. First, we plan to study and implement model compression techniques such as distillation to reduce the size of all the components in our proposed framework and implement the compressed model on physical device considering the resource constraints. Second, we will seek to work on federated learning using the proposed framework to protect user's data from leaving her device and build a robust framework considering well-known attacks to federated learning. Finally, for PII type detection, we will work upon class-incremental learning to address new leak types that appear after model training.
\bibliographystyle{./IEEEtran}
\bibliography{./main}

\end{document}